%% file: iclr2024_conference.tex
\documentclass{article} 
\usepackage{iclr2024_conference,times}

\input{math_commands.tex}

\usepackage{hyperref}
\usepackage{url}
\usepackage{microtype}
\usepackage{times}
\usepackage{CJKutf8}
\usepackage{xcolor}
\usepackage{latexsym}
\usepackage{tcolorbox}
\usepackage{multirow}
\usepackage{tikz}
\usepackage{capt-of}
\usepackage{graphicx}  
\usepackage{pgfplots}
\pgfplotsset{compat=1.12}
\usepackage{amsmath}
\usepackage{multicol}
\usepackage{booktabs}
\usepackage{color}
\usepackage{mwe}
\usepackage{wrapfig}
\usepackage{colortbl,array,xcolor}
\usepackage{xspace}

\newcommand{\model}{\textsc{Self-Rag}\xspace}
\newcommand{\mgen}{$\mathcal{M}$\xspace}
\newcommand{\mcrt}{$\mathcal{C}$\xspace}
\newcommand{\mret}{$\mathcal{R}$\xspace}
\usepackage{tikz}
\usetikzlibrary{tikzmark}
\makeatletter
\newcommand*\myfontsize{%
  \@setfontsize\myfontsize{6.7}{8}%
}
\makeatother
\newcommand{\mytextbox}[2]{\tikzmarknode[draw=#1,thick,inner sep=2pt]{test}{\myfontsize #2}}

\definecolor{cadmiumgreen}{rgb}{0.0, 0.42, 0.24}

\definecolor{myred}{rgb}{0.7, 0.3, 0.0}
\definecolor{myblue}{rgb}{0.2, 0.3, 0.6}
\newcommand{\ret}{\mytextbox{myred}{\textbf{\textcolor{myred}{Retrieve}}}}
\newcommand{\crt}{\mytextbox{myblue}{\textbf{\textcolor{myblue}{Critique}}}}

\newcommand{\crel}{
    \mytextbox{myblue}{
        \textbf{\textcolor{myblue}{\textsc{IsRel}}}
    }
}
\newcommand{\cgr}{
    \mytextbox{myblue}{
        \textbf{\textcolor{myblue}{\textsc{IsSup}}}
    }
}
\newcommand{\cuse}{
    \mytextbox{myblue}{
        \textbf{\textcolor{myblue}{\textsc{IsUse}}}
    }
}

\newcolumntype{R}[1]{>{\raggedleft\let\newline\\\arraybackslash\hspace{0pt}}m{#1}}
\usetikzlibrary{intersections}

\definecolor{darkgreen}{rgb}{0.0, 0.42, 0.24}
\usepackage{caption}
\usepackage{subcaption}
\usepackage{graphicx}
\usepackage{pifont}
\usepackage{titletoc}

\usepackage{amsfonts}
\usepackage{booktabs}
\usepackage{arydshln}
\usepackage{colortbl}
\usepackage{algorithm}
\usepackage[noend]{algpseudocode}
\usepackage{enumitem}
\usepackage{graphicx}
\usepackage{soul}
\usepackage{colortbl,array,xcolor}

\title{Self-Rag: Self-reflective Retrieval augmented Generation}
\title{\textsc{Self-rag:} Learning to Retrieve, Generate, and Critique through Self-Reflection}

\author{Akari Asai$^\dagger$, Zeqiu Wu$^\dagger$, Yizhong Wang$^{\dagger\S}$, Avirup Sil$^\ddagger$, Hannaneh Hajishirzi$^{\dagger\S}$ \\
$^\dagger$University of Washington~~~~~$^\S$Allen Institute for AI~~~~~$^\ddagger$IBM Research AI \\
\texttt{\{akari,zeqiuwu,yizhongw,hannaneh\}@cs.washington.edu}, \texttt{avi@us.ibm.com} \\
}

\usepackage{color, colortbl}
\definecolor{azure}{rgb}{0.0, 0.5, 1.0}

\iclrfinalcopy
\begin{document}

\maketitle

\begin{abstract}
Despite their remarkable capabilities, large language models (LLMs) often produce responses containing factual inaccuracies due to their sole reliance on the parametric knowledge they encapsulate. 
Retrieval-Augmented Generation (RAG), an ad hoc approach that augments LMs with retrieval of relevant knowledge, decreases such issues. 
However, indiscriminately retrieving and incorporating a fixed number of retrieved passages, regardless of whether retrieval is necessary, or passages are relevant, diminishes LM versatility or can lead to unhelpful response generation.
We introduce a new framework called {\bf Self-Reflective Retrieval-Augmented Generation (\model)} that enhances an LM's quality and factuality through retrieval and self-reflection. 
Our framework trains a single arbitrary LM that adaptively retrieves passages on-demand, and generates and reflects on retrieved passages and its own generations using special tokens, called {\it reflection} tokens. Generating reflection tokens makes the LM controllable during the inference phase, enabling it to tailor its behavior to diverse task requirements. 
Experiments show that \model (7B and 13B parameters) significantly outperforms state-of-the-art LLMs and retrieval-augmented models on a diverse set of tasks. 
Specifically, \model outperforms ChatGPT and retrieval-augmented Llama2-chat on Open-domain QA, reasoning and fact verification tasks, and it shows significant gains in improving factuality and citation accuracy for long-form generations relative to these models.\footnote{{Our code and trained models are available at \url{https://selfrag.github.io/}}.}

\end{abstract}

\section{Introduction}
\input{sections/intro}

\section{Related Work}
\input{sections/related}

\section{\model: Learning to Retrieve, Generate and Critique}
\input{sections/method_v3}

\section{Experiments}
\input{sections/experiments}

\section{Results and Analysis}
\input{sections/analysis}

\section{Conclusion}
\input{sections/conclusion}


\subsubsection*{Acknowledgments}
We thank Sewon Min, Scott Wen-tau Yih, Sean Welleck, and Kawin Ethayarajh for fruitful discussions in the early stages of this work. 
We thank Sewon Min, Joongwon (Daniel) Kim, and Sandy Kaplan for valuable feedback on the paper, and Tianyu Gao and Weijia Shi for their help on evaluations. 
Akari Asai is supported by the IBM Fellowship. 
We thank Stability AI for providing computing to train and evaluate the LMs in this work, and Microsoft Accelerate Foundation Models Research Program for the access to OpenAI APIs. 
This work was funded in part by the DARPA MCS program through NIWC Pacific (N66001-19-2-4031), NSF IIS-2044660, and gifts from AI2.

\bibliography{iclr2024_conference}
\bibliographystyle{iclr2024_conference}

\clearpage
\appendix
\section*{Appendix}
\input{sections/appendix}

\end{document}

%% file: math_commands.tex
\usepackage{amsmath,amsfonts,bm}

\def\eqref#1{equation~\ref{#1}}

\def\1{\bm{1}}

\DeclareMathAlphabet{\mathsfit}{\encodingdefault}{\sfdefault}{m}{sl}
\SetMathAlphabet{\mathsfit}{bold}{\encodingdefault}{\sfdefault}{bx}{n}



%% file: sections/intro.tex
\begin{figure}[t!]
\includegraphics[width=\textwidth]{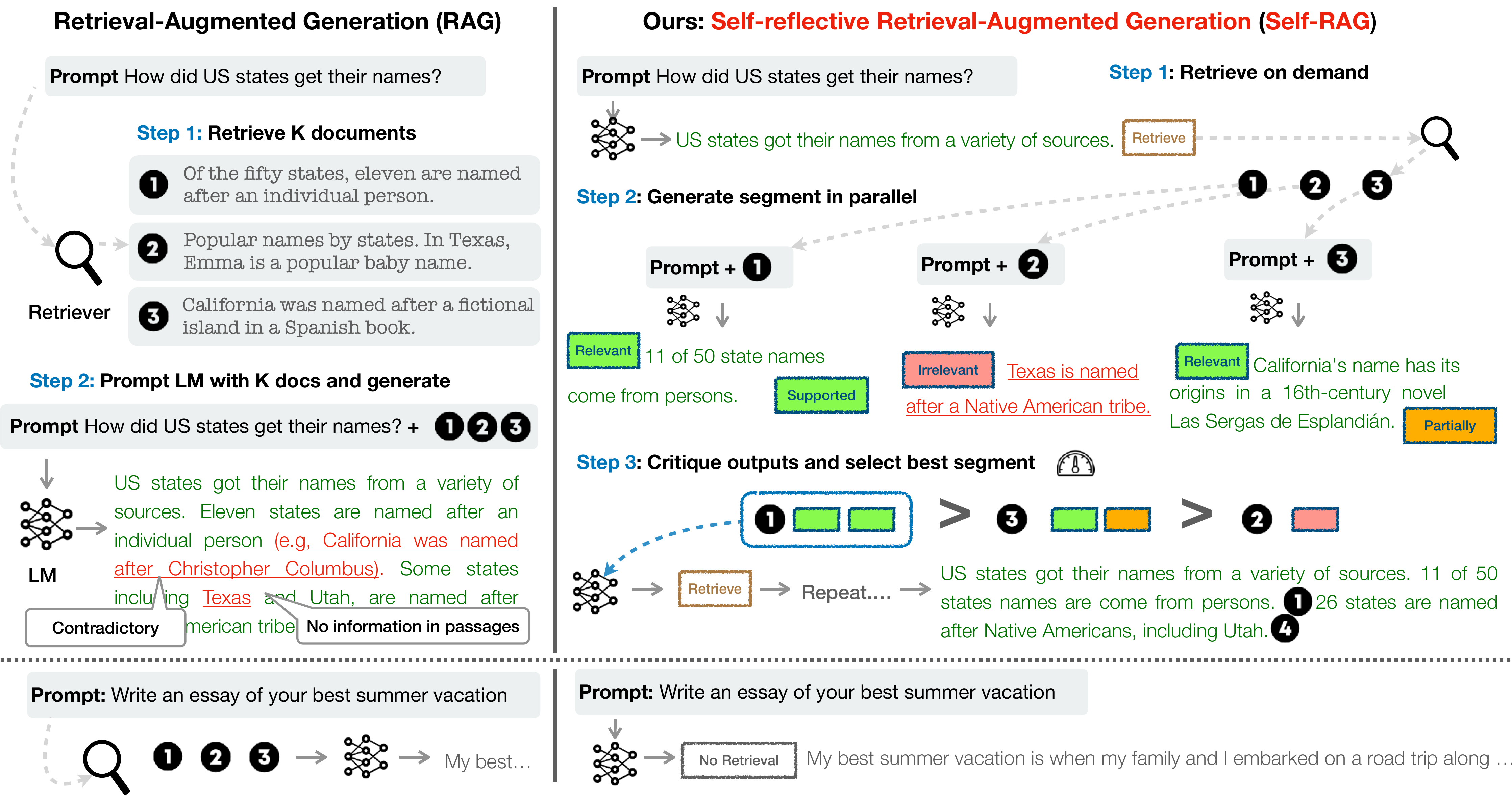}\caption{Overview of \model. \model learns to retrieve, critique, and generate text passages to enhance overall generation quality, factuality, and verifiability. 
} \label{fig:taxonomy}
\end{figure}

State-of-the-art LLMs continue to struggle with factual errors~\citep{mallen2022not,min2023factscore} despite their increased model and data scale~\citep{ouyang2022training}. 
Retrieval-Augmented Generation (RAG) methods (Figure~\ref{fig:taxonomy} left;~\citealt{lewis2020retrieval,guu2020retrieval}) augment the input of LLMs with relevant retrieved passages,  reducing factual errors in knowledge-intensive tasks~\citep{ram2023context, asai-etal-2023-retrieval}. 
However, these methods may hinder the versatility of LLMs or introduce unnecessary or off-topic passages that lead to low-quality generations ~\citep{pmlr-v202-shi23a} since they retrieve passages indiscriminately regardless of whether the factual grounding is helpful.
Moreover, the output is not guaranteed to be consistent with retrieved relevant passages~\citep{gao2023enabling} since the models are not explicitly trained to leverage and follow facts from provided passages. 
This work introduces {\bf Self-Reflective Retrieval-augmented Generation (\model)} to improve an LLM's generation quality, including its factual accuracy without hurting its versatility, via on-demand retrieval and self-reflection. 
We train an arbitrary LM in an end-to-end manner to learn to reflect on its own generation process given a task input by generating both task output and intermittent special tokens (i.e., \textit{reflection tokens}). Reflection tokens are categorized into \textit{retrieval} and \textit{critique} tokens to indicate the need for retrieval and its generation quality respectively (Figure~\ref{fig:taxonomy} right). 
{
In particular, given an input prompt and preceding generations, \model first determines if augmenting the continued generation with retrieved passages would be helpful. If so, 
it outputs a {\bf retrieval} token that calls a retriever model on demand (Step 1). 
Subsequently, \model concurrently processes multiple retrieved passages, evaluating their relevance and then {\bf generating} corresponding task outputs (Step 2). It then generates critique tokens to \textbf{criticize} its own output and choose best one (Step 3) in terms of factuality and overall quality. 
This process differs from conventional RAG (Figure~\ref{fig:taxonomy} left), which consistently retrieves a fixed number of documents for generation regardless of the retrieval necessity (e.g., the bottom figure example does not require factual knowledge) and never second visits the generation quality.
Moreover, \model provides citations for each segment with its self-assessment of whether the output is supported by the passage, leading to easier fact verification. }

\model trains an arbitrary LM to generate text with reflection tokens by unifying them as the next token prediction from the expanded model vocabulary. 
{
We train our generator LM on a diverse collection of text interleaved with reflection tokens and retrieved passages. 
Reflection tokens, inspired by reward models used in reinforcement learning~\citep{ziegler2019fine, ouyang2022training}, are inserted offline into the original corpus by a trained {\it critic} model. This eliminates the need to host a critic model during training, reducing overhead.
The critic model, in part, is supervised on a dataset of input, output, and corresponding reflection tokens collected by prompting a propriety LM (i.e., GPT-4; \citealt{openai2023gpt}).
While we draw inspiration from studies that use control tokens 
to start and guide text generation~\citep{lu2022quark,keskar2019ctrl},
our trained LM uses critique tokens to assess its own predictions after each generated segment as an integral part of the generation output.
}

{\model further enables a customizable 
decoding algorithm to satisfy hard or soft constraints, which are defined by reflection token predictions.}
In particular, our inference-time algorithm enables us to (1) flexibly adjust retrieval frequency for different downstream applications and (2) customize models' behaviors to user preferences by leveraging reflection tokens through segment-level beam search using the weighted linear sum of the reflection token probabilities as segment score.

Empirical results on six tasks, including reasoning and long-form generation, demonstrate that \model significantly outperforms pre-trained and instruction-tuned LLMs that have more parameters and  widely adopted RAG approaches with higher citation accuracy. 
In particular, \model outperforms retrieval-augmented ChatGPT on four tasks, Llama2-chat~\citep{touvron2023llama} and Alpaca~\citep{dubois2023alpacafarm} on all tasks. 
Our analysis demonstrates the effectiveness of training and inference with reflection tokens for overall performance improvements as well as test-time model customizations (e.g., balancing the trade-off between citation previsions and completeness).

%% file: sections/related.tex
\noindent{\bf Retrieval-Augmented Generation.}
{
Retrieval-Augmented Generation (RAG) augments the input space of LMs with retrieved text passages~\citep{guu2020retrieval,lewis2020retrieval}, leading to large improvements in knowledge-intensive tasks after fine-tuning or used with off-the-shelf LMs~\citep{ram2023context}. 
A more recent work~\citep{luo2023sail} instruction-tunes an LM with a fixed number of retrieved passages prepended to input, or pre-train a retriever and LM jointly, followed by few-shot fine-tuning on task datasets~\citep{izacard2022few}. }
While prior work often retrieves only once at the beginning, \cite{jiang2023active} propose to
adaptively retrieve passages for generation on top of a proprietary LLM or ~\cite{schick2023toolformer} train an LM to generate API calls for named entities. 
Yet, the improved task performance of such approaches often comes at the expense of runtime efficiency~\citep{mallen2022not}, robustness to irrelevant context~\citep{pmlr-v202-shi23a}, and lack of attributions~\citep{liu2023evaluating,gao2023enabling}. 
We introduce a method to train an arbitrary LM to learn to use retrieval \textit{on-demand} for diverse instruction-following queries and introduce controlled generation guided by reflections tokens to further improve generation quality and attributions. 

\noindent{\bf Concurrent RAG work. }
{
A few concurrent works\footnote{All work is arXived within a week of this preprint.} on RAG propose new training or prompting strategies to improve widely-adopted RAG approaches. 
\cite{lin2023radit} fine-tune both the retriever and LM on instruction-tuning datasets in two steps.
While we also train our model on diverse instruction-following datasets, \model enables retrieval on demand and selection of the best possible model output via fine-grained self-reflection, making it widely applicable and more robust and controllable. 
\citet{Yoran2023MakingRL} use a natural language inference model and \citet{xu2023recomp} use a summarization model to filter out or compress retrieved passages before using them to prompt the LM to generate the output.
\model processes passages in parallel and filters out irrelevant ones through self-reflection, without relying on external models at inference. Moreover, our self-reflection mechanism also evaluates other aspects of the model output quality including factuality.
LATS~\citep{zhou2023language} prompt off-the-shelf LMs to search for relevant information for question answering tasks and to generate with tree search, guided by LM-generated value scores. 
While their value function simply indicates an overall score of each generation, \model trains to an arbitrary LM to learn to generate fine-grained self-reflection and customizable inference. 
}


\noindent{\bf Training and generating with critics.}
Training LLMs with reinforcement learning (e.g., Proximal Policy Optimization or PPO; \citealt{schulman2017proximal}) from human feedback (RLHF) has proven effective in aligning LLMs with human preferences~\citep{ouyang2022training}. 
\citet{wu2023fine} introduce fine-grained RLHF with multiple reward models. 
Though our work also studies fine-grained critique on retrieval and generation, we train our target LM on task examples augmented with reflection tokens from a critic 
model offline, with a far lower training cost compared to RLHF. In addition, reflection tokens in \model enable controllable generation at inference, while RLHF focuses on human preference alignment during training.
Other works use general control tokens to guide LM generation~\citep{lu2022quark,korbak2023pretraining}, while \model uses reflection tokens to decide the need for retrieval and to self-evaluate generation quality.
\cite{xie2023decomposition} propose a self-evaluation-guided decoding framework, but they focus only on reasoning tasks with one evaluation dimension (reasoning path consistency) and without retrieval.
%
Recent work on LLM refinement \citep{dhuliawala2023chainofverification, madaan2023selfrefine, paul2023refiner} prompts a model to generate task output, natural language feedback and refined task output iteratively, but at the cost of inference efficiency.

%% file: sections/method_v3.tex
We introduce Self-Reflective Retrieval-Augmented Generation (\model), shown in Figure~\ref{fig:taxonomy}. \model is a framework that enhances the quality and factuality of an LLM through retrieval and self-reflection, without sacrificing LLM's original creativity and versatility. 
Our end-to-end training lets an LM \mgen ~{\bf generate} text informed by {\bf retrieved} passages, if needed, and {\bf criticize} the output by learning to generate special tokens. 
These {\it reflection tokens} (Table~\ref{tab:reward}) signal the need for retrieval or confirm the output's relevance, support, or completeness. 
In contrast, common RAG approaches retrieve passages indiscriminately, without ensuring complete support from cited sources. 

\subsection{Problem Formalization and Overview}
\label{sec:overview}
Formally, given input $x$, we train \mgen to sequentially generate textual outputs $y$ consisting of multiple segments $y=[y_1, \dots, y_T]$, where $y_t$ indicates a sequence of tokens for the $t$-th segment.\footnote{
In this paper, we treat one sentence as a segment in our experiments, but our framework is applicable to any segment unit (i.e., sub-sentence).} 
Generated tokens in $y_t$ include text from the original vocabulary as well as the reflection tokens (Table~\ref{tab:reward}). 

\begin{table*}[t!]
\centering
\footnotesize
\begin{tabular}{p{2cm}p{1.2cm}p{3.5cm}p{5.5cm}}\toprule
Type & Input & Output & Definitions \\ \midrule
\ret   & $x$ / $x,y$ & \{yes, no, continue\} & Decides when to retrieve with \mret \\
\crel  & $x, d$  &\{{\bf relevant}, irrelevant\}  &   $d$ provides useful information to solve $x$. \\
\cgr  & $x, d, y$  &\{{\bf fully supported}, partially supported, no support\}  & All of the verification-worthy statement in $y$ is supported by $d$.   \\
\cuse & $x, y$  &\{{\bf 5}, 4, 3, 2, 1\}  & $y$ is a useful response to $x$.   \\
\bottomrule
\end{tabular}
    \caption{Four types of reflection tokens used in \model. Each type uses several tokens to represent its output values. 
    The bottom three rows are three types of \crt~tokens, and {\bf the bold text} indicates the most desirable critique tokens. $x, y, d$ indicate input, output, and a relevant passage, respectively. 
    }
    \label{tab:reward}
\end{table*}

\begin{algorithm}
\caption{\model Inference}\label{alg:self_reward inference}
\begin{algorithmic}[1]
\Require Generator LM \mgen, Retriever \mret, Large-scale passage collections $\{d_1, \ldots, d_N\}$ 
\State {\bf Input:} input prompt $x$ and preceding generation $y_{<t}$, {\bf Output:} next output segment $y_t$
\State \mgen predicts \ret~given $(x,y_{<t})$
\If{\ret~ == \texttt{Yes}}
    \State Retrieve relevant text passages $\mathbf{D}$  using \mret given $(x,y_{t-1})$ \Comment{\textcolor{blue}{Retrieve} }
    \State \mgen predicts \crel given $x, d$ and $y_{t}$ given $x,d, y_{<t}$ for each $d \in \mathbf{D}$ \Comment{\textcolor{blue}{Generate} }
    \State \mgen predicts \cgr and \cuse given $x, y_t, d$ for each $d \in \mathbf{D}$ \Comment{\textcolor{blue}{Critique} }
    \State Rank $y_t$ based on \crel, \cgr, \cuse \Comment{\textcolor{gray}{Detailed in Section~\ref{sec:fine_grained_decoding} }}
\ElsIf{\ret~ == \texttt{No}}
    \State $\mathcal{M}_{gen}$ predicts $y_t$ given $x$ \Comment{\textcolor{blue}{Generate} }
    \State $\mathcal{M}_{gen}$ predicts \cuse given $x,y_t$\Comment{\textcolor{blue}{Critique} }
\EndIf
\end{algorithmic}\label{algo:inference}
\end{algorithm}

\noindent {\bf Inference overview.}
Figure~\ref{fig:taxonomy} and Algorithm~\ref{algo:inference} present an overview of \model at inference. 
For every $x$ and preceding generation $y_{<t}$, the model decodes a retrieval token to evaluate the utility of retrieval. If retrieval is not required, the model predicts the next output segment, as it does in a standard LM.   
If retrieval is needed, the model generates: a critique token to evaluate the retrieved passage's relevance, the next response segment, and a critique token to evaluate if the information in the response segment is supported by the passage. Finally, a  new critique token evaluates the overall utility of the response.\footnote{We follow \citet{liu2023evaluating} in using a ``perceived'' utility value that is independent of retrieved passages.} 
To generate each segment, \model processes multiple passages in parallel and uses its own generated reflection tokens to enforce soft constraints (Section~\ref{sec:fine_grained_decoding}) or hard control (Algorithm~\ref{algo:inference}) over the generated task output. 
For instance, in Figure~\ref{fig:taxonomy} (right), the retrieved passages $d_1$ is selected at the first time step since $d_2$ does not provide direct evidence (\crel is {Irrelevant}) and $d_3$ output is only partially supported while $d_1$ are fully supported. 

\noindent {\bf Training overview.} 
\model enables an arbitrary LM to generate text with reflection tokens by unifying them as next token predictions from the expanded model vocabulary (i.e., the original vocabulary plus reflection tokens). Specifically, we train the generator model \mgen on a curated corpus with interleaving passages retrieved by a {\it retriever} \mret and reflection tokens predicted by a {\it critic} model \mcrt (summarized in Appendix  Algorithm~\ref{alg:self_reward training}). We train \mcrt to generate reflection tokens for evaluating retrieved passages and the quality of a given task output (Section \ref{sec:critic_trainig}). 
Using the critic model, we update the training corpus by inserting reflection tokens into task outputs offline.  
Subsequently, we train the final generator model (\mgen) using the conventional LM objective (Section \ref{sec:gen_training}) to enable \mgen to generate reflection tokens by itself without relying on the critic at inference time. 

\subsection{\model Training}
Here,  we describe the supervised data collection and training of two models, the critic \mcrt 
(Section~\ref{sec:critic_trainig}) and the generator \mgen (Section~\ref{sec:gen_training}).

\subsubsection{Training the Critic Model}
\label{sec:critic_trainig}

\noindent{\bf Data collection for critic model.} 
Manual annotation of reflection tokens for each segment is expensive~\citep{wu2023fine}. 
A state-of-the-art LLM like GPT-4~\citep{openai2023gpt} can be effectively used to generate such feedback~\citep{liu2023gpteval}. 
However, depending on such proprietary LMs can raise API costs and diminish reproducibility
~\citep{chen2023chatgpt}. 
We create supervised data by prompting GPT-4 to generate reflection tokens and then distill their knowledge into an in-house \mcrt. 
For each group of reflection tokens, we randomly sample instances from the original training data: $\{X^{sample}, Y^{sample}\} \sim \{X, Y\}$. 
{
As different reflection token groups have their own definitions and input, as shown in Table~\ref{tab:reward}, we use different instruction prompts for them. Here, we use \ret~ as an example. We prompt GPT-4 with a type-specific instruction (``Given an instruction, make a judgment on whether finding some external documents from the web helps to generate a better response.'') followed by few-shot demonstrations $I$ the original task input $x$ and output ${y}$ to predict an appropriate reflection token as text: $p(r|I, x, y)$.
Manual assessment reveals that GPT-4 reflection token predictions show high agreement with human evaluations.
We collect 4k-20k supervised training data for each type and combine them to form training data for \mcrt. 
Appendix Section~\ref{sec:instruction_gpt_4} shows the full list of instructions, and \ref{sec:data_collection_gpt4}} contains more details and our analysis.

\paragraph{Critic learning.}
After we collect training data $\mathcal{D}_{critic}$, we initialize \mcrt with a pre-trained LM 
and train it on $\mathcal{D}_{critic}$ using a standard conditional language modeling objective, maximizing likelihood:
\begin{equation}\label{eq:crt_training}
    \max_{\mathcal{C}}\mathbb{E}_{((x,y),r) \sim \mathcal{D}_{critic}} \log p_{\mathcal{C}} (r| x, y), \text{ 
$r$ for reflection tokens. }
\end{equation}
Though the initial model can be any pre-trained LM, we use the same one as the generator LM (i.e., Llama 2-7B; \citealt{touvron2023llama}) for \mcrt initialization. 
The critic achieves a higher than 90\% agreement with GPT-4-based predictions on most reflection token categories (Appendix Table~\ref{tab:reward_prediction}). 

\begin{figure}[t!]
  \begin{center}
    \includegraphics[width=0.95\textwidth]{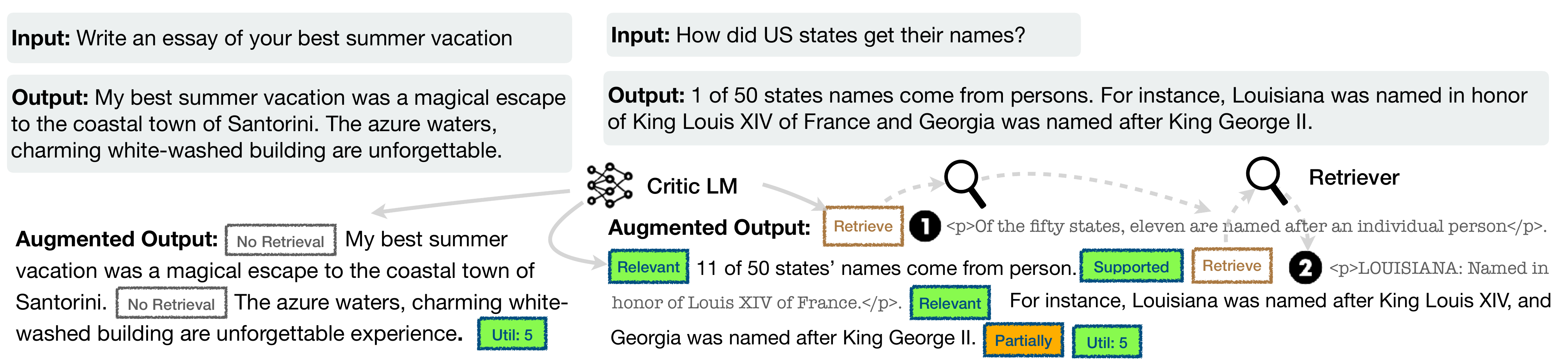}
  \end{center}
  \caption{\model training examples. The left example does not require retrieval while the right one requires retrieval; thus, passages are inserted. More examples are in Appendix Table~\ref{tab:examplse_training_table}.}\label{fig:training_example}
\end{figure}

\subsubsection{Training the Generator Model}
\label{sec:gen_training}

\paragraph{Data collection for generator.}
Given an input-output pair $(x, y)$, we augment the original output $y$ using the retrieval and critic models to create supervised data that precisely mimics the \model inference-time process (Section~\ref{sec:overview}). 
For each segment $y_t \in y$, we run \mcrt to assess whether additional passages could help to enhance generation.  
If retrieval is required, the retrieval special token \ret~=\texttt{Yes} is added, and \mret retrieves the top $K$ passages, $\mathbf{D}$. 
For each passage, \mcrt further evaluates whether the passage is relevant and predicts \crel. 
If a passage is relevant, \mcrt further evaluates whether the passage supports the model generation and predicts \cgr. 
Critique tokens \crel and \cgr are appended after the retrieved passage or generations.  
At the end of the output, $y$ (or $y_T$),  \mcrt predicts the overall utility token \cuse 
, and an augmented output with reflection tokens and the original input pair is added to $\mathcal{D}_{gen}$. 
See the example training data in Figure~\ref{fig:training_example}.

\noindent{\bf Generator learning.}
We train the generator model \mgen by training on the curated corpus augmented with reflection tokens $\mathcal{D}_{gen}$ using the standard next token objective:
\begin{equation}\label{eq:gen_training}
    \max_{\mathcal{M}}\mathbb{E}_{(x,y,r) \sim \mathcal{D}_{gen}} \log p_{\mathcal{M}} (y, r| x). 
\end{equation}
Unlike \mcrt training (Eq.~\ref{eq:crt_training}), \mgen learns to predict the target output as well as the reflection tokens.  
During training, we mask out the retrieved text chunks (surrounded by \texttt{<p>} and \texttt{</p>} in Figure~\ref{fig:training_example}) for loss calculation and 
expand the original vocabulary $\mathcal{V}$ with a set of reflection tokens $\{\crt, \ret\}$. 

\paragraph{Connections to prior work on learning with critique.}
Recent work 
incorporates additional critique (feedback) during training, e.g.,  RLHF~(\citealt{ouyang2022training}) via PPO. 
While PPO relies on separate reward models during training, we compute critique offline and directly insert them into the training corpus, where the generator LM is trained with a standard LM objective. This significantly reduces training costs compared to PPO. 
Our work also relates to prior work that incorporates special tokens to control generation ~\citep{keskar2019ctrl,lu2022quark, korbak2023pretraining}. Our \model learns to generate special tokens {\it to evaluate its own prediction} after each generated segment,  
enabling the use of a soft re-ranking mechanism or hard constraints at inference (discussed next). 

\subsection{\model Inference}
\label{sec:fine_grained_decoding}
Generating reflection tokens to self-evaluate its own output makes \model controllable during the inference phase, enabling it to tailor its behavior to diverse task requirements. 
For tasks demanding factual accuracy~\citep{min2023factscore}, we aim for the model to retrieve passages more frequently to ensure that the output aligns closely with the available evidence. Conversely, in more open-ended tasks, like composing a personal experience essay, the emphasis shifts towards retrieving less and prioritizing the overall creativity or utility score. In this section, we describe approaches to enforce control to meet these distinct objectives during the inference process.

\noindent{\bf Adaptive retrieval with threshold.}
\model dynamically decides when to retrieve text passages by predicting \ret. 
Alternatively, our framework allows a threshold to be set. Specifically, if the probability of generating the {\ret=\texttt{Yes}} token normalized over all output tokens in \ret~ surpasses a designated threshold, we trigger retrieval (details in Appendix Section~\ref{sec:inference_details}).

\noindent{\bf Tree-decoding with critique tokens.}
At each segment step $t$, when retrieval is required,  based either on hard or soft conditions, \mret retrieves $K$ passages, and the generator 
\mgen processes each passage in parallel and outputs $K$ different continuation candidates. 
We conduct a segment-level beam search (with the beam size=$B$) to obtain the top-$B$ segment continuations at each timestamp $t$, and return the best sequence at the end of generation. The score of each segment $y_t$ with respect to passage $d$ is updated with a critic score $\mathcal{S}$
that is the linear weighted sum of the normalized probability of each \crt~token type. 
For each critique token group $G$ (e.g., \crel), we denote its score at timestamp $t$ as $s_t^G$, 
and we compute a segment score as follows:  
\begin{equation}
    f(y_t,  d, \crt) = p(y_{t} | x, d, y_{<t})) + \mathcal{S}(\crt){\rm , where}
\end{equation}
\begin{equation}\label{eq:reward}
    \mathcal{S}(\crt) = \sum_{G \in \mathcal{G}} w^G s_t^G 
    \mbox{ for } \mathcal{G} = \{\crel, \cgr, \cuse\}, 
\end{equation}
where  $s_t^G = \frac{p_t(\hat{r})}{\sum_{i=1}^{N^G} p_t(r_i)}$ stands for the generation probability of the most desirable reflection token $\hat{r}$ (e.g., \crel=\texttt{Relevant}) for the critique token type $G$ with $N^G$ distinct tokens (that represent different possible values for $G$). 
The weights $w^G$ in Eq.~\ref{eq:reward} are hyperparameters that can be adjusted at inference time to enable customized behaviors at test time. 
For instance, to ensure that result $y$ is mostly supported by evidence, we can set a weight term for the \cgr score higher, while relatively lowering weights for other aspects. 
Alternatively, we could further enforce hard constraints during decoding using \crt. 
Instead of using a soft reward function in Eq.~\ref{eq:reward}, we could explicitly filter out a segment continuation when the model generates an undesirable $\crt$ token (e.g., \cgr=\texttt{No support}) . 
{Balancing the trade-off between multiple preferences has been studied in RLHF~\citep{touvron2023llama,wu2023fine}, which often requires training to change models' behaviors. \model tailors an LM with no additional training.  
}

%% file: sections/experiments.tex
\subsection{Tasks and Datasets} 
We conduct evaluations of our \model and diverse baselines on a range of downstream tasks, holistically evaluating outputs with metrics designed to assess overall correctness, factuality, and fluency. 
Throughout these experiments, we conduct zero-shot evaluations, where we provide instructions describing tasks without few-shot demonstrations~\citep{wei2022finetuned,sanh2022multitask}. 
Details of our experiments' settings, including test-time instructions, are available in the Appendix Section~\ref{sec:training_details}.

\noindent{\bf Closed-set tasks} include two datasets, i.e., a fact \textit{verification dataset} about public health ({\bf PubHealth}; \citealt{zhang2023interpretable}) and a \textit{multiple-choice reasoning dataset} created from scientific exams ({\bf ARC-Challenge}; ~\citealt{clark2018think}). 
We use accuracy as an evaluation metric and report on the test set. 
We aggregate the answer probabilities of target classes for both of these datasets (Appendix Section~\ref{sec:details_experiments}).

\noindent{\bf Short-form generations tasks} include two open-domain question answering (QA) datasets, PopQA~\citep{mallen2022not} and TriviaQA-unfiltered~\citep{joshi-etal-2017-triviaqa}, where systems need to answer arbitrary questions about factual knowledge. 
For PopQA, we use the long-tail subset, consisting of 1,399 rare entity queries whose monthly Wikipedia page views are less than 100. 
As the TriviaQA-unfiltered (open) test set is not publicly available, we follow prior work's validation and test split~\citep{min-etal-2019-discrete,guu2020retrieval}, using 11,313 test queries for evaluation. 
We evaluate performance based on whether gold answers are included in the model generations instead of strictly requiring exact matching, following \cite{mallen2022not,schick2023toolformer}. 

\noindent{\bf Long-form generation tasks} include a biography generation task~\citep{min2023factscore} and a long-form QA task~{\bf ALCE-ASQA}~\cite{gao2023enabling,stelmakh2022asqa}. 
We use FactScore~\citep{min2023factscore} to evaluate biographies, and we use official metrics of correctness (str-em), fluency based on MAUVE~\citep{pillutla2021MAUVE}, and citation precision and recall~\citep{gao2023enabling} for ASQA. \footnote{\url{https://github.com/princeton-nlp/ALCE}} 

\subsection{Baselines}
\noindent{\bf Baselines without retrievals.}
We evaluate strong publicly available pre-trained LLMs, Llama2$_{\textsc{7b},\textsc{13b}}$~\citep{touvron2023llama}, instruction-tuned models, Alpaca$_{\textsc{7b},\textsc{13b}}$~\citep{dubois2023alpacafarm} (our replication based on Llama2); and models trained and reinforced using private data, ChatGPT~\citep{ouyang2022training} and Llama2-chat$_{\textsc{13b}}$. 
For instruction-tuned LMs, we use the official system prompt or instruction format used during training if publicly available. 
We also compare our method to concurrent work, CoVE$_\textsc{65b}$~\citep{dhuliawala2023chainofverification}, which introduces iterative prompt engineering to improve the factuality of LLM generations. 

\noindent{\bf Baselines with retrievals.}  
We evaluate models augmented with retrieval at test time or during training. The first category includes standard RAG baselines, where an LM (Llama2, Alpaca) generates output given the query prepended with the top retrieved documents using the same retriever as in our system. It also includes {Llama2-FT}, where Llama2 is fine-tuned on all training data we use without the reflection tokens or retrieved passages.
{We also report the result of retrieval-augmented baselines with LMs trained with private data: Ret-ChatGPT and Ret-Llama2-chat, which deploy the same augmentation technique above, as well as perplexity.ai, an InstructGPT-based production search system. 
}
The second category includes concurrent methods that are trained with retrieved text passages, i.e., { SAIL}~\citep{luo2023sail} to instruction-tune an LM on the Alpaca instruction-tuning data with top retrieved documents inserted before instructions, and {Toolformer}~\citep{schick2023toolformer} to pre-train an LM with API calls (e.g., Wikipedia APIs).\footnote{We report numbers using the results reported in the paper as the implementations are not available.}

\subsection{Experimental settings}
\noindent{\bf Training data and settings.}
Our training data consists of diverse instruction-following input-output pairs. 
In particular, we sample instances from Open-Instruct processed data~\citep{wang2023far} and knowledge-intensive datasets~\citep{petroni-etal-2021-kilt,stelmakh2022asqa,mihaylov-etal-2018-suit}. 
In total, we use 150k instruction-output pairs. 
We use Llama2 7B and 13B~\citep{touvron2023llama} as our generator base LM, and we use Llama2 7B as our base critic LM. 
For the retriever model \mret, we use off-the-shelf Contriever-MS MARCO~\citep{izacard2021towards} by default and retrieve up to ten documents for each input. 
More training details are in the Appendix Section~\ref{sec:training_details}. 

\noindent{\bf Inference settings.}
As a default configuration, we assign the weight terms \crel, \cgr, \cuse values of 1.0, 1.0 and 0.5, respectively.
 To encourage frequent retrieval, we set the retrieval threshold to 0.2 for most tasks and to 0 for ALCE~\citep{gao2023enabling} due to citation requirements.    
We speed up inference using vllm~\citep{kwon2023efficient}. 
At each segment level, we adopt a beam width of 2.  
For a token-level generation, we use greedy decoding.  
By default, we use the top five documents from Contriever-MS MARCO~\citep{izacard2021towards}; for biographies and open-domain QA, we use additional top five documents retrieved by a web search engine, following~\cite{luo2023sail}; for ASQA, we use the author-provided top 5 documents by GTR-XXL~\citep{ni2021large} across all baselines for a fair comparison.

%% file: sections/analysis.tex
\begin{table*}[t!]
\centering
\footnotesize
    \caption{
    Overall experiment results on six tasks. {\bf Bold} numbers indicate the best performance among non-proprietary models, and \textcolor{gray}{\bf gray-colored} bold text indicates the best proprietary model when they outperforms all non-proprietary models.  
    $^*$ indicates concurrent or recent results reported by concurrent work.  -- indicates numbers that are not reported by the original papers or are not applicable. Models are sorted based on scale. FS, em, rg, mau, prec, rec denote FactScore (factuality); str-em, rouge (correctness); MAUVE (fluency); citation precision and recall, respectively. 
    }
    \label{tab:main}
\begin{tabular}{lrrrrrrrrrr}\toprule
&\multicolumn{2}{c}{Short-form} & \multicolumn{2}{c}{Closed-set} & \multicolumn{6}{c}{Long-form generations (with citations)}     \\
& PopQA & TQA &  Pub & ARC &  Bio  &\multicolumn{5}{c}{ASQA} \\ 
LM & (acc) & (acc) & (acc) & (acc) & (FS) & (em) & (rg) & (mau) & (pre) & (rec) \\
\midrule
\multicolumn{11}{c}{\it LMs with proprietary data } \\
Llama2-c$_{\textsc{13b}}$  & 20.0 & 59.3 & 49.4 & 38.4 & 55.9 & 22.4 & 29.6 & 28.6 & -- & -- \\
Ret-Llama2-c$_{\textsc{13b}}$  & 51.8  & 59.8& 52.1 & 37.9 &  {79.9} & 32.8 & 34.8 & 43.8 & 19.8& 36.1 \\
ChatGPT  & 29.3 &   \textcolor{gray}{\bf 74.3} & 70.1 &  \textcolor{gray}{\bf 75.3}  & 71.8 & 35.3  & 36.2  & 68.8 & -- & --\\ 
Ret-ChatGPT &  50.8 & 65.7 &  54.7 & \textcolor{gray}{\bf 75.3} & -- & \textcolor{gray}{\bf 40.7} &  \textcolor{gray}{\bf 39.9} & \textcolor{gray}{\bf 79.7} & {65.1} & \textcolor{gray}{\bf 76.6}  \\
Perplexity.ai &  -- & -- & -- & -- & 71.2 & -- &  -- & -- & -- &--  \\ \midrule
\multicolumn{11}{c}{\it Baselines without retrieval} \\
Llama2$_{\textsc{7b}}$ & 14.7 & 30.5 & 34.2 & 21.8  & 44.5  & 7.9 & 15.3&19.0 & -- & --  \\
Alpaca$_{\textsc{7b}}$  &  23.6 & 54.5  & 49.8 &45.0 & 45.8 & 18.8 & 29.4 & 61.7 & -- & --  \\
Llama2$_{\textsc{13b}}$   &  14.7 & 38.5  & 29.4 & 29.4 & 53.4 & 7.2 & 12.4 & 16.0  & -- & -- \\
Alpaca$_{\textsc{13b}}$ &  24.4  &  61.3  & 55.5  & 54.9 & 50.2 & 22.9 &  32.0&  70.6 & -- & -- \\
CoVE$_{\textsc{65b}}$ * &  -- & -- & -- & -- & 71.2  & --  & -- & -- & -- & -- \\  
\midrule

\multicolumn{11}{c}{\it Baselines with retrieval} \\
Toolformer*$_{\textsc{6b}}$  & -- & 48.8  & -- & -- & -- & --  & -- & -- & -- & --  \\ 
Llama2$_{\textsc{7b}}$  & 38.2  & 42.5 & 30.0 & 48.0 & 78.0 & 15.2 & 22.1 & 32.0 & 2.9 & 4.0 \\
Alpaca$_{\textsc{7b}}$   & 46.7 & 64.1  & 40.2 & 48.0 & 76.6  & 30.9 & 33.3 & 57.9 & 5.5 & 7.2 \\
Llama2-FT$_{\textsc{7b}}$  & 48.7 & 57.3 & 64.3 & 65.8& {78.2} & 31.0 & 35.8 & 51.2 & 5.0 & 7.5 \\
SAIL*$_{\textsc{7b}}$  &  -- & -- & 69.2 & 48.4 & -- & --  & -- & -- & -- & --  \\ 
Llama2$_{\textsc{13b}}$  &  45.7  & 47.0 & 30.2  & 26.0 & 77.5  & 16.3 & 20.5 & 24.7 & 2.3 & 3.6  \\
Alpaca$_{\textsc{13b}}$ & 46.1& {66.9}  &  51.1 & 57.6 & 77.7 & {\bf 34.8} & 36.7 & 56.6 & 2.0 & 3.8 \\
\hdashline
{\bf Our }\model$_{\textsc{7b}}$ &{  54.9} &  66.4  & { 72.4} & 67.3 & {\bf 81.2} & 30.0 & 35.7 & {\bf 74.3} & {66.9} & { 67.8} \\
{\bf Our }\model$_{\textsc{13b}}$  &  {\bf 55.8} & {\bf 69.3} & {\bf 74.5} & {\bf 73.1} &  80.2 & 31.7 & {\bf 37.0} & { 71.6} & {\bf 70.3} & {\bf 71.3} \\
\bottomrule
\end{tabular}
\end{table*}

\subsection{Main Results} 
\noindent{\bf Comparison against baselines without retrieval. }
Table~\ref{tab:main} (top) presents the baselines without retrieval. 
Our \model (bottom two rows) demonstrates a substantial performance advantage over supervised fine-tuned LLMs in all tasks and even outperforms ChatGPT in PubHealth, PopQA, biography generations, and ASQA (Rouge and MAUVE). 
Our approach also significantly outperforms a concurrent method that employs sophisticated prompt engineering; specifically, on the bio generation task, our 7B and 13B models outperform the concurrent CoVE~\citep{dhuliawala2023chainofverification}, which iteratively prompts Llama2$_\textsc{65b}$ to refine output. 

\noindent{\bf Comparison against baselines with retrieval.}
As shown in Tables~\ref{tab:main} (bottom), our \model also outperforms existing RAG in many tasks, obtaining the best performance among non-proprietary LM-based models on all tasks. 
While our method outperforms other baselines, on PopQA or Bio, powerful instruction-tuned LMs with retrieval (e.g., LLama2-chat, Alpaca) show large gains from their non-retrieval baselines. 
However, we found that these baselines provide limited solutions for tasks where we cannot simply copy or extract sub-strings of retrieved passages. 
On PubHealth and ARC-Challenge, baselines with retrieval do not improve performance notably from their no-retrieval counterparts.
We also observe that most baselines with retrieval struggle to improve citation accuracy. 
On ASQA, our model shows significantly higher citation precision and recall than all models except ChatGPT. 
\citet{gao2023enabling} found that ChatGPT consistently exhibits superior efficacy in this particular task, surpassing smaller LMs. 
{
Our \model bridges this performance gap, even outperforming ChatGPT in citation precision, which measures whether the model-generated claim is fully supported by cited evidence. } 
We also found that on the metrics for factual precision, \model 7B occasionally outperforms our 13B due to the tendency of smaller \model to often generate precisely grounded yet shorter outputs. 
{
Llama2-FT$_{\textsc{7b}}$, which is the baseline LM trained on the same instruction-output pairs as \model without retrieval or self-reflection and is retrieval-augmented at test time only, lags behind \model. This result indicates \model gains are not solely from training data and demonstrate the effectiveness of \model framework.  }

\begin{figure}[t!]
\begin{subfigure}[b]{0.38\textwidth}
\centering
\footnotesize
\begin{tabular}{lccc}
\toprule
 & PQA & Med & AS \\
 & (acc) & (acc) & (em) \\
\midrule
\model (50k) & 45.5 & 73.5  & 32.1 \\ \hdashline
\multicolumn{4}{l}{\it Training } \\
No Retriever \mret & 43.6 & 67.8 & 31.0 \\
No Critic \mcrt &  42.6 & 72.0 & 18.1  \\\hdashline
\multicolumn{4}{l}{\it Test } \\
No retrieval & 24.7 &  73.0 & --  \\
Hard constraints&  28.3 &  72.6 & --  \\
Retrieve top1 & 41.8 & 73.1 & 28.6  \\
Remove \cgr & 44.1  & 73.2 & 30.6 \\ 
\bottomrule
\end{tabular}
\caption{ Ablation}
\label{tab:ablatioon}
\end{subfigure}%
  \hspace{.3cm}
  \begin{subfigure}[b]{0.265\textwidth}
        \includegraphics[width=\textwidth]{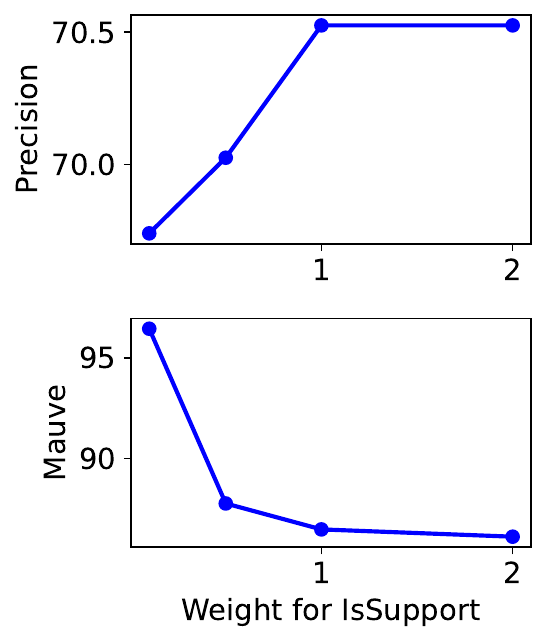}
        \caption{Customization}
        \label{fig:customization}
  \end{subfigure}%
    \begin{subfigure}[b]{0.325\textwidth}
        \includegraphics[width=\textwidth]{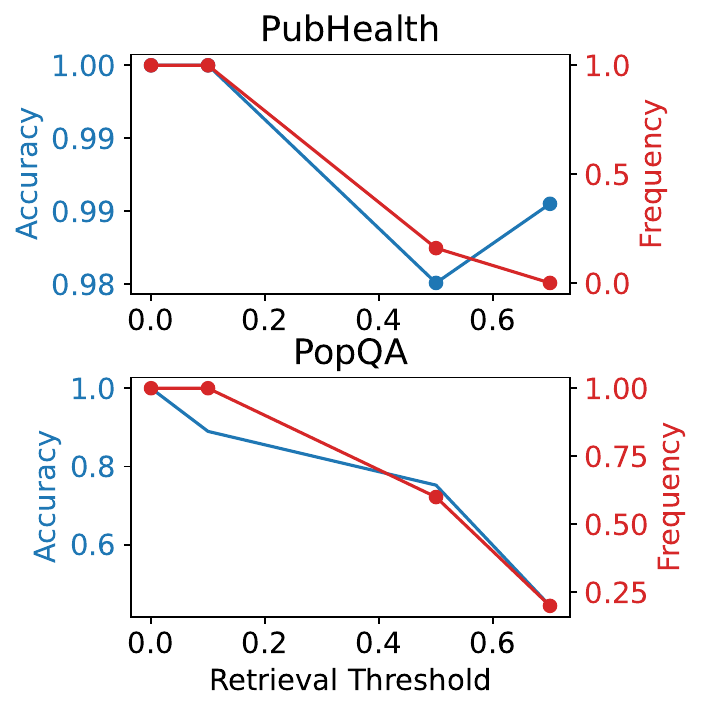}
        \caption{Retrieval}
        \label{fig:retrieval_freq}
  \end{subfigure}%
\caption{{\bf Analysis on \model:} (a) {\bf Ablation studies} for key components of \model training and inference based on our 7B model.  (b) {\bf Effects of soft weights} on ASQA citation precision and Mauve (fluency). (c) {\bf Retrieval frequency} and {\it normalized} accuracy on PubHealth and PopQA.   }
\end{figure}
\subsection{Analysis}
\label{sec:analyis}
\paragraph{Ablation studies.}
We conduct a set of ablations of our framework to identify which factors play key roles. 
We evaluate two model variants trained differently than our model: {\it No Retriever} trains an LM using the standard instruction-following method given instruction-output pairs, without retrieved passages;  {\it No Critic} trains an LM trained with input-output pairs that are always augmented with the top one retrieved document without reflection tokens. 
This is similar to SAIL~\citep{luo2023sail}, and we use our instruction-output data instead of using the Alpaca dataset~\citep{dubois2023alpacafarm}, as in SAIL. 
We also conduct ablation on our inference-time algorithm, including {\it No retrieval} disables retrieval during inference; {\it Hard constraints} indicates the model performance that retrieves when \ret=\texttt{Yes} instead of using the adaptive threshold; {\it Retrieve top 1} always retrieves and uses the top one document only, similar to standard RAG approaches; {\it Remove \cgr} indicates the model performance that removes \cgr score only during critique-guided beam search in Eq.~\ref{eq:reward}. 
{In this ablation experiment, we use a training instance size of 50k for a more efficient exploration of training variations. Later in this section, we conduct an analysis of the effect of training data size. 
We conduct the ablation studies on three datasets, PopQA, PubHealth, and ASQA. On ASQA, we evaluate models on sampled 150 instances and exclude ablations involving adaptive or no retrieval processes. }

We show in Table~\ref{tab:ablatioon} the ablation results. 
The top part of the table shows results for training ablations, and the bottom part is for inference ablations. We see that all components play important roles.
We also observe a large performance gap between \model and No Retriever or Critic baselines across tasks, indicating that training an LM with those models largely contributes to the performance gain of \model. 
{
Using the top passages regardless of their relevance (Retrieve top 1) as in conventional RAG approaches causes a large drop in PopQA and ASQA, and removing \cgr during the beam search results hurts performance on ASQA. 
This demonstrates the effectiveness of \model's capabilities of carefully selecting generations based fine-grained multiple criterion, instead of naively using all of the top passages from the retrieval model or solely depending on relevance scores. 
}

\noindent{\bf Effects of inference-time customization.}
One key benefit of our proposed framework is that it enables us to control how much each critique type affects the final generation sampling. We analyze the effects of different parameter weights on the top of our 7B model during inference time on ASQA, where multiple evaluation aspects are considered. 
Figure~\ref{fig:customization} shows the effects of changing the weighting term for \cgr, which criticizes how supported the output is by the text passage. 
As the figure shows, increasing the weight leads to positive effects on the models' citation precision since this puts more emphasis on whether model generation is supported by the evidence. 
{
On the contrary, a larger weight results in lower MAUVE scores: when generation gets longer and more fluent, there are often more claims that are not fully supported by citations, consistent with findings by \cite{liu2023evaluating}. }
Our framework lets practitioners choose and customize models' behaviors at test time by adjusting such parameters without requiring additional training. 

\noindent{\bf Efficiency and accuracy trade-off.}
Using our framework, practitioners can adjust how often retrieval occurs using the token probability of reward tokens. We evaluate how this adaptive threshold affects overall accuracy and frequency of retrieval, and we evaluate the performance with varying numbers of threshold $\delta$ (larger $\delta$ results in less retrieval) on PubHealth and PopQA. 
Figure~\ref{fig:retrieval_freq} shows that the model's retrieval frequencies dramatically change on both datasets. as $\delta$ varies.  
On one hand, performance deterioration by retrieving less is smaller on PubHealth but larger in PopQA. 

\begin{figure}[t!]
  \begin{subfigure}[b]{0.24\textwidth}
        \includegraphics[width=\textwidth]{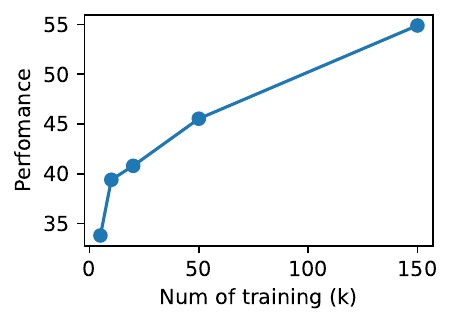}
        \caption{PopQA}
        \label{fig:popqa_scale}
  \end{subfigure}%
    \begin{subfigure}[b]{0.21\textwidth}
        \includegraphics[width=\textwidth]{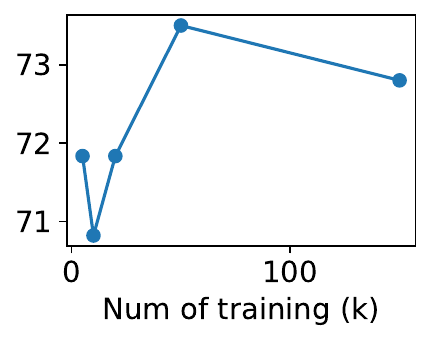}
        \caption{PubHealth}
        \label{fig:med_sacle}
  \end{subfigure}%
    \begin{subfigure}[b]{0.21\textwidth}
        \includegraphics[width=\textwidth]{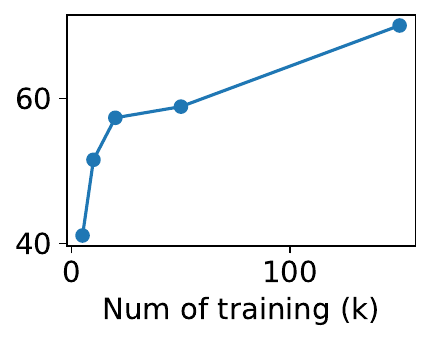}
        \caption{ASQA (prec)}
        \label{fig:asqa_prec_scaling}
  \end{subfigure}
\begin{subfigure}[b]{0.3\textwidth}
\centering
\footnotesize
\begin{tabular}{lcc}
\toprule
& Pop & Bio. \\
\midrule
S \& P & 92.5 & 70.0 \\ \hdashline
\crel & 95.0 & 90.0 \\
\cgr & 90.0 &  85.0 \\
\bottomrule
\end{tabular}
\caption{Human evaluation on PopQA and Bio generation. }
\label{tab:human_eval}
\end{subfigure}%
  \hspace{.3cm}
\caption{{\bf Training scale and Human analysis:} (a) (b) (c) {\bf Training scale analysis} shows the effect of the training data scale on PopQA, PubHealth and ASQA (citation precision), respectively.  
(d) {\bf Human analysis} on \model outputs as well as reflection tokens. }
\end{figure}

\noindent{\bf Effects of training data size.}
{We conduct an analysis of how the data scale affects the model's performance. In particular, we randomly sample 5k, 10k, 20k, and 50k instances from our original 150k training instances, and fine-tune four \model$_{\textsc{7b}}$ variants on those subsets. Then, we compare the model performance on PopQA, PubHealth, and ASQA (citation precision) with our final \model trained on the full 150k instances. 
We also evaluate 
Figures~\ref{fig:popqa_scale}, \ref{fig:med_sacle} and \ref{fig:asqa_prec_scaling} shows the models' performance trained on different amount of data. 
Across all datasets, increasing data size often shows upward trajectories and the improvements are significantly larger in PopQA and ASQA, while we do not observed such significant improvements on Llama2-FT$_{\textsc{7b}}$ when increasing the training data from 50k to 150k. 
These results also indicate that further expanding the training data of \model may lead to further improvements, although in this work we limit our training data size to 150k. 
}

\noindent{\bf Human evaluations.}
{We conduct small human evaluations on \model outputs, as well as the reliability of predicted reflection tokens. 
In particular, we sampled 50 samples from PopQA and Bio results. 
Following \citet{menick2022teaching}, human annotators evaluate {\it S\&P}, which indicates whether the model output is plausible (i.e., the output is a reasonable and on-topic response to the question as if it were occurring in a conversation) and supported (i.e., the provided evidence is sufficient to verify
the validity of the answer). For {S\&P}, we do not consider the instances where \model predicts \texttt{irrelevant} or \texttt{no support}. 
We then ask our annotators whether the model-predicted reflection tokens about \crel and \cgr match their inspections (e.g., whether the {\it fully supported} output is supported by the cited evidence).  
Human annotators find \model answers are often plausible and supported by relevant passages with higher S\&P scores on short-form PopQA, which is consistent with \citet{menick2022teaching}. 
Human annotators also find \crel and \cgr reflection token predictions are mostly aligned with their assessments. 
Appendix Table~\ref{tab:human_evaluations} shows several annotated examples and explanations on assessments. 
}

%% file: sections/conclusion.tex
This work introduces \model, a new framework to enhance the quality and factuality of LLMs through retrieval on demand and self-reflection. 
\model trains an LM to learn to retrieve, generate, and critique text passages and its own generation by predicting the next tokens from its original vocabulary as well as newly added special tokens, called reflection tokens. 
\model further enables the tailoring of LM behaviors at test time by leveraging reflection tokens. 
Our holistic evaluations on six tasks using multiple metrics demonstrate that \model significantly outperforms LLMs with more parameters or with conventional retrieval-augmented generation approaches. 

\section*{Ethical Concerns}
This work aims to improve the factuality of LLM outputs, the lack of which continues to cause numerous real-world problems (e.g., spread of misinformation and provision of incorrect and dangerous advice). While our method shows significant improvements in terms of performance, factuality, and citation accuracy, it can still generate outputs that are not fully supported by the citations. 
We hope that explicit self-reflection and fine-grained attribution may help users verify factual errors in the model outputs.

%% file: sections/appendix.tex
\startcontents[sections]
\printcontents[sections]{l}{1}{\setcounter{tocdepth}{2}}
\newpage

\section{\model Details} 
\subsection{Reflection Tokens.}
\label{sec:data_collection_gpt4}
\paragraph{Definitions of reflection tokens.} Below, we provide a detailed definition of reflection type and output tokens.   
The first three aspects will be provided at each segment level, while the final aspect is only given at each output level.
\begin{itemize}[leftmargin=*]
    \item {\bf Retrieval-on-demand} (\ret): Given an input and previous-step generation (if applicable), an LM determines whether the continuation requires factual grounding. 
    \texttt{No} indicates retrieval is unnecessary as the sequence does not require factual grounding or may not be enhanced by knowledge retrieval, \texttt{Yes} indicates retrieval is necessary. 
    We additionally have \texttt{continue to use evidence}, which indicates that a model can continue to use the evidence retrieved previously. For instance, a passage may contain rich factual information, and thus \model generates multiple segments based on the passage. 
    \item {\bf Relevant} (\crel): Retrieved knowledge may not be always relevant to the input. This aspect indicates whether the evidence provides useful information (\texttt{Relevant}) or not (\texttt{Irrelevant}). 
    \item {\bf Supported} (\cgr): Attribution is the concept of whether the output is fully supported by certain evidence~\citep{menick2022teaching,bohnet2022attributed}. 
    This aspect judges how much information in the output is entailed by the evidence. 
    We evaluate attributions in three scale, \texttt{Fully supported}, \texttt{Partially supported}, and \texttt{No support / Contradictory}, following \cite{yue2023automatic,nakano2021webgpt}. 
    \item {\bf Useful} (\cuse): Following the definitions from ~\cite{liu2023evaluating}, we define the perceived utility as whether the response is a helpful and informative answer to the query, independently from whether it is in fact factual or not. This can be also viewed as plausibility in \citet{menick2022teaching}. For usefulness, we use a five-scale evaluation (1 is the lowest and 5 is the highest).
\end{itemize}

\paragraph{Details of GPT-4-based data collections.} 
We use the instruction and demonstration pairs to prompt GPT-4, listed in Section~\ref{sec:instruction_gpt_4}. 
Following an official recommendation, we separate instructions and outputs with ``\#\#''. 
We use the temperature 1 and set the maximum output token counts to be 200. 
We discard instances where GPT-4 does not follow the designated output formats or output sequences that do not match our expected category names. 
As a result, we collected 1,2594 for \ret, 11,181 for \cgr, 19,317 for relevance, 3,831 for utility.  

\paragraph{Manual analysis of the GPT-4 predictions.}
The authors of this paper manually assess randomly sampled 20 instances for each aspect and check if GPT-4 predictions match their assessments given the same instruction, demonstrations, and test instances. 
We found our assessments show high agreement with GPT-4 predictions, especially for relevance (95\%), retrieval necessity (95\%), and the degree of support (90\%). Agreement was slightly lower in usefulness (80\%), mostly due to the disagreement between 1 and 2 or 4 and 5.   

\subsection{\model Training}
\paragraph{Overview of training.}
Algorithm~\ref{alg:self_reward training} provides a high-level overview of our training.

\begin{algorithm}
\caption{\model Training}\label{alg:self_reward training}
\begin{algorithmic}[1]
\State {\bf Input} input-output data $\mathcal{D} = \{X, Y\}$, generator $\mathcal{M}$, \mcrt $\theta$
\State Initialize \mcrt with a pre-trained LM
\State Sample data $\{X^{sample}, Y^{sample}\} \sim \{X, Y\}$ \Comment{\textcolor{blue}{\bf Training Critic LM (Section~\ref{sec:critic_trainig}) }}
\For{$(x, y) \in (X^{sample}, Y^{sample})$} \Comment{\textcolor{blue}{Data collections for \mcrt} }
\State Prompt GPT-4 to collect a reflection token $r$ for $(x,y)$ 
\State Add $\{(x, y, r)\}$ to $\mathcal{D}_{critic}$ 
\EndFor
\State Update \mcrt with next token prediction loss \Comment{\textcolor{blue}{ Critic learning; Eq.~\ref{eq:crt_training} }}
\State Initialize \mgen with a pre-trained LM \Comment{\textcolor{blue}{\bf Training Generator LM (Section~\ref{sec:gen_training}) }}
\For{$(x, y) \in (X, Y)$} \Comment{\textcolor{blue}{Data collection for \mgen with $\mathcal{D}_{critic}$ } }
\State Run \mcrt to predict $r$ given $(x, y)$
\State Add $(x, y, r)$ to $\mathcal{D}_{gen}$ 
\EndFor
\State Update \mgen on $\mathcal{D}_{gen}$ with next token prediction loss \Comment{\textcolor{blue}{ Generator LM learning}; Eq.~\ref{eq:gen_training} }
\end{algorithmic}
\end{algorithm}

\paragraph{Full list of seed datasets.}
To sample diverse input-output pairs, we sample instances of the Open-Instruct~\citep{wang2023far} dataset. In particular, we use their ShareGPT, GPT-4 Alpaca, Alpaca, OpenAssistant, and FLAN subsets subsets. 
We also sample instances from a couple of knowledge-intensive datasets, Natural Questions~\citep{kwiatkowski-etal-2019-natural}, Wizard of Wikipedia~\citep{dinan2018wizard} and FEVER~\citep{thorne-etal-2018-fever} from the KILT benchmark~\citep{petroni-etal-2021-kilt}, ASQA~\citep{stelmakh2022asqa} and multiple QA datasets including ARC-Easy and OpenBookQA~\citep{mihaylov-etal-2018-suit}. 
Table~\ref{tab:mgen_training} shows the full list of training instances, and in total, we use {145,619} instances. 
\begin{table}[t!]
\centering
\small
\begin{tabular}{l|c c c}
\toprule
Dataset name & category & Data source & the number of instances  \\
\midrule
GPT-4 Alpaca &  Instruction-following & Open-Instruct & 26,168\\ 
Stanford Alpaca &  Instruction-following & Open-Instruct & 25,153 \\ 
FLAN-V2 &  Instruction-following & Open-Instruct &17,817 \\
ShareGPT & Instruction-following & Open-Instruct & 13,406 \\
Open Assistant 1 &  Instruction-following & Open-Instruct & 9,464 \\
Wizard of Wikipedia &  Knowledge-intensive & KILT & 17,367  \\
Natural Questions &  Knowledge-intensive & KILT & 15,535\\ 
FEVER &  Knowledge-intensive & KILT & 9,966\\ 
OpenBoookQA &  Knowledge-intensive & HF Dataset & 4,699 \\ 
Arc-Easy &  Knowledge-intensive & HF Dataset & 2,147 \\ 
ASQA &  Knowledge-intensive & ASQA &3,897 \\ 
\bottomrule
\end{tabular}
\caption{The generator LM \mgen training data statistics.}\label{tab:mgen_training}
\end{table}

\paragraph{Performance of the Critic \mcrt.}
We evaluate the accuracy of reward predictions by splitting GPT-4 generated feedback into training, development, and test sets. 
The accuracy of the reward model is as follows. Table~\ref{tab:reward_prediction} shows the model performance of predicting GPT-4 judgments. 
As you can see, overall our fine-tuned reward model shows high prediction matching with GPT-4 predicted feedback. 
While our final model uses Llama2-7B as a base LM, we also train and compare FLAN-3B~\citep{wei2022finetuned} model on the same data, to investigate the effectiveness of different data sizes affect final reward predictions. 
In most aspects, our reward model shows higher than 80\% accuracy, indicating the powerful ability of fine-tuned specialized LMs to evaluate text. While both models show relatively lower performance on \cuse, this is because both models often confuse between the two highest cases (5 and 4), where human annotators can also disagree.   
\begin{figure}[t!]
\begin{minipage}{\textwidth}
\centering
\small
\begin{tabular}{l|cccc}
\toprule
base LM & \ret & \cgr & \crel & \cuse \\
\midrule
Llama2-7B & {\bf 93.8} & {\bf 93.5} &  80.2  & {\bf 73.5}  \\
FLAN-3B & 85.6 & 73.1 & {\bf 82.0} & 72.1  \\
\bottomrule
\end{tabular}
\caption{Reward prediction accuracy using GPT-4 predictions as ground-truth predictions. 
}\label{tab:reward_prediction}
\end{minipage}
\end{figure}

\paragraph{Details of \mgen data creation.}
 Here, we provide detailed data creation procedures. 
Algorithm~\ref{algo:data_aug} summarizes the process. 
Here we set $y_t$ to $y$ for simplification.   
Once we train the critic model, we first run it on input data from the aforementioned datasets, to predict whether retrieval is needed or not. For the instances where the critic predicts \ret=\texttt{No}, we only predict the \cuse given input and output. For the instances where the critic predicts \ret=\texttt{Yes}, we first retrieve passages using the input and the entire output as queries, to find passages that are relevant to the entire output. We then split output sentences using Spacy.\footnote{\url{https://spacy.io/}} 
For each sentence, we run \mcrt to predict whether the retrieval is necessary or not, given the input, preceding segments, and the initial retrieved passage.  
If \mcrt predicts \ret=\texttt{No}, then do not insert any paragraph at the $t$th segment. 
If \mcrt predicts \ret=\texttt{Yes}, then we use the original input and the $t$th segment as a retrieval query to find relevant passages for the $t$-th segment. For each retrieved passage, we predict \crel and \cgr. 
If there is any passage and continuation with \crel=\texttt{Relevant} and \cgr=\texttt{Fully Supported} / \cgr=\texttt{Partially Supported}, then we sample it as the continuation. If there is more than one passage satisfying this criterion, we use the one with the highest retrieval score. 
If there are only \crel=\texttt{Irrelevant} or \cgr=\texttt{No Support} passages, we randomly sample one passage. 

\begin{algorithm}
\caption{$\mathcal{M}_{gen}$ Data creation}\label{algo:data_aug}
\begin{algorithmic}[1]
\State {\bf Input}  Input-output data $\mathcal{D}$ = ${X, Y}$
\For{$(x, y) \in \{X, Y\}$} 
\State Given $(x,y)$ \mcrt predicts \ret
\If{\ret~is predicted}
    \State Retrieve relevant passages $\mathbf{D}$ using \mret given $(x,y)$
    \Comment{\textcolor{blue}{Retrieve passages} }
    \For {$d \in \mathbf{D}$}
    \State \mcrt predicts \crel for each $d$     \Comment{\textcolor{blue}{Predict relevance of passages} }
     \State \mcrt predicts \cgr for each $(y, d)$ \Comment{\textcolor{blue}{Predict supports of outputs} }
    \EndFor
    \State\mcrt predicts \cuse for each $d$ \Comment{\textcolor{blue}{Predict overall utility ($t=\mathcal{T}$ only)} }
    \State Sample $d$
\ElsIf{\ret~is not predicted}
    \State \mcrt predicts \cuse given $x, y$
\EndIf
Add augmented $(x, y, d, r)$ to $\mathcal{D}_{gen}$
\EndFor
\end{algorithmic}
\end{algorithm}

\paragraph{Training examples.}
Table~\ref{tab:examplse_training_table} show several training examples used for \mgen training.

\begin{table*}[t!]
\renewcommand{\arraystretch}{1.2}
\setlength{\tabcolsep}{2pt}
\footnotesize
    \centering
    \begin{tabular}{p{14cm}}
\toprule
{\bf Input} Identify the two types of computer memory. (from \texttt{gpt4\_alpaca})\\
{\bf Output} [\ret=\texttt{Yes}]\texttt{<p>}\textcolor{gray}{Computer memory Computer memory -- In computing, memory refers to the computer hardware integrated circuits that store information for immediate use in a computer; it is synonymous with the term ``primary storage''. Computer memory operates at a high speed, for example random-access memory (RAM), as a distinction from storage that provides slow-to-access information but offers higher capacities. If needed, contents of the computer memory can be transferred to secondary storage; a very common way of doing this is through a memory management technique called ``virtual memory''. An archaic synonym for memory is store. The term ``memory'', meaning ``primary storage'' or ``main memory'', is}\texttt{<p>}[\crel=\texttt{Relevant}]Primary memory, also known as main memory, is the memory that the computer can directly access and manipulate. [\cgr=\texttt{Fully Supported}] [\ret=\texttt{Yes}]\texttt{<p>}\textcolor{gray}{Computer memory--often associated with addressable semiconductor memory, i.e. integrated circuits consisting of silicon-based transistors, used for example as primary storage but also other purposes in computers and other digital electronic devices. There are two main kinds of semiconductor memory, volatile and non-volatile. Examples of non-volatile memory are flash memory (used as secondary memory) and ROM, PROM, EPROM and EEPROM memory (used for storing firmware such as BIOS). Examples of volatile memory are primary storage, which is typically dynamic random-access memory (DRAM), and fast CPU cache memory, which is typically static random-access memory (SRAM) that is fast but energy-consuming, offering lower memory}\texttt{<p>}[\crel=\texttt{Relevant}]Primary memory can be accessed quickly, but it is volatile, meaning that the data is lost if the power is turned off.[\cgr=\texttt{Partially Supported}]...[\cuse:5] \\ \midrule
{\bf Input} Give an example of a situation where you have to delegate tasks (from \texttt{gpt4\_alpaca})\\
{\bf Output} [\ret=\texttt{No}]The manager realizes that there is an extensive list of tasks that need to be completed in order to make the event a success.[\ret=\texttt{Yes}]\texttt{<p>}\textcolor{gray}{Delegation--Delegation is the assignment of any responsibility or authority to another person (normally from a manager to a subordinate) to carry out specific activities. It is one of the core concepts of management leadership. However, the person who delegated the work remains accountable for the outcome of the delegated work. Delegation empowers a subordinate to make decisions, i.e. it is a shifting of decision-making authority from one organizational level to a lower one. Delegation, if properly done, is not fabrication. The opposite of effective delegation is micromanagement, where a manager provides too much input,}\texttt{<p>}[\crel=\texttt{Relevant}]This includes tasks such as creating a guest list, designing and sending invitations, booking a venue, coordinating with vendors, planning the event\'s program, and monitoring the budget.[\cgr=\texttt{Partially Supported}][\ret=\texttt{No}]Since it is not possible for the manager to handle everything alone, they must delegate tasks to their team members.[\ret=\texttt{No}]The manager would then assign each team member specific responsibilities based on their individual skillsets and strengths.[\ret=\texttt{Yes}]\textcolor{gray}{\texttt{<p>}Joint application design -- Subject Matter Experts: These are the business users, the IS professionals, and the outside experts that will be needed for a successful workshop. This group is the backbone of the meeting\texttt{<p>}}[\crel=\texttt{Irrelevant}]For instance, the team\'s graphic designer might be tasked with designing the invitations, while the event coordinator would be in charge of planning the program and working with vendors...[\ret=\texttt{No}]By delegating tasks, the manager can ensure that all necessary tasks are completed in a timely manner, while also empowering their team members to take on responsibilities and contribute to the success of the event.[\cuse=5]\\
\midrule
{\bf Input} Synthesize a poem based on this prompt:``The sun is setting'' (from \texttt{stanford\_alpaca})\\
{\bf Output} [\ret=\texttt{No}]The stars begin to twinkle bright the trees, scattered throughout the night; [\ret=\texttt{No}]The stars begin to twinkle bright the trees, scattered throughout the night [\ret=\texttt{No}] They flicker like a silent dance, A peaceful moment in its trance. [\ret=\texttt{No}] They flicker like a silent dance, A peaceful moment in its trance. [\ret=\texttt{No}] A chorus of crickets sing a lullaby, To every creature that roams the sky [\cuse=5]\\ \midrule
{\bf Input} when did walking dead season 7 come out  (from \texttt{nq})\\
{\bf Output} [\ret=\texttt{Yes}]\texttt{<p>}\textcolor{gray}{The Walking Dead (season 7)--The Walking Dead (season 7) The seventh season of ``The Walking Dead'', an American post-apocalyptic horror television series on AMC, premiered on October 23, 2016, and concluded on April 2, 2017, consisting of 16 episodes. Developed for television by Frank Darabont, the series is based on the eponymous series of comic books by Robert Kirkman, Tony Moore, and Charlie Adlard. ...}\texttt{<p>}[\crel=\texttt{Relevant}]October 23 , 2016[\cgr=\texttt{Fully Supported}][\cuse=5] \\ \midrule
{\bf Input} Given a chat history separated by new lines, generates an informative, knowledgeable and engaging response: Hello, I love German shepherd dogs and you? -- My favorite dog,  they are officially known as German Shepherd Dogs? -- Yes, I have a German shepherd in my house and she is pregnant (from \texttt{wow})\\
{\bf Output} hey also are easily trained, obedient, and intelligent.[\ret=\texttt{Yes}]\texttt{<p>}\textcolor{gray}{German Shepherd -- a simple difference of opinion, it is the fundamental issue of the breed\'s essential conformation and movement. ... They are known for being easy to train and good for performing tasks and following instructions. They are especially well known for their police work, being used for tracking criminals, patrolling troubled areas and detecting and holding suspects. }\texttt{<p>}[\crel=\texttt{Relevant}]Because of this, they are often used for disability assistance and for police work. [\cgr=\texttt{Partially Supported}][\cuse=2] \\ 
\bottomrule
 \end{tabular}
    \caption{List of the training examples.}\label{tab:examplse_training_table}
\end{table*}

\subsection{\model Inference}
\label{sec:inference_details}
\paragraph{Details of beam-search score calculations.}
We first compute scores for each critique type by taking the normalized probabilities of desirable tokens. 
For \crel, we compute the score as follows:
\begin{equation*}
    s(\crel) = \frac{p(\crel = \textsc{Relevant})}{p(\crel = \textsc{Relevant}) + p(\crel = \textsc{Irrelevant})}.
\end{equation*}
For \cgr, we compute the score as follows:
\begin{equation*}
    s(\crel) = \frac{p(\cgr = \textsc{Fully})}{S} + 0.5 \times \frac{p(\cgr = \textsc{Partially})}{S}, 
\end{equation*}
where $S = \sum_{t \in \{\textsc{Fully}, \textsc{Partially}, \textsc{No}\}}p(\cgr=t)$. 
For \cuse where we have a five-scale score, we compute the weighted sum of the scores. 
We assigns weighted scores of $w =\{-1, -0.5, 0, 0.5, 1\}$ to the tokens \cuse=$\{1, 2, 3, 4, 5\}$, and compute the final scores as follows:
\begin{equation*}
    s(\cuse) = \sum_{i}^5 w_i\frac{p(\cuse = i)}{S},
\end{equation*}
where $S =\sum_{t \in \{1,2,3,4,5\}}p(\cuse=t)$. 
\paragraph{Details of adaptive retrieval.}
For retrieval based on soft constraints, we trigger retrieval if the following condition is satisfied: 
\begin{equation*}
   \frac{p(\ret = \textsc{Yes})}{p(\ret = \textsc{Yes})+ p(p(\ret = \textsc{No})} > \delta.
\end{equation*}
\section{Experimental Details}


\subsection{More Details of Training }
\label{sec:training_details}
\paragraph{More details of training and computations.}
We use 4 Nvidia A100 with 80GB memory to train our models. All models are trained for 3 epochs with a batch size of 128, a peak learning rate of 2e-5 with 3\% warmup steps, and linear decay afterward. We set the maximum token length to be 2,048 for the 7B model, and 1,524 for the 13B model due to the memory constraint. We use Deepspeed stage 3~\citep{rajbhandari2020zero} to conduct multi-GPU distributed training, with training precision Bfloat16 enabled. FlashAttention~\citep{dao2022flashattention} is used to make the long-context training more efficient.
We run inference of our trained models using 1-2 Quadro RTX 6000 GPUs with 24GB memory. 

\subsection{More Details of Evaluations}
\label{sec:details_experiments}
\paragraph{Retrieval setup details.}
By default, we use Contriever-MS MARCO to retrieve the top five documents from Wikipedia{, and use official Wikipedia embeddings based on 2018 English Wikipedia. On PopQA, where question and answer pairs are created based on WikiData in 2022, we found that the 2018 Wikipedia sometimes lacks articles about some entities that have been more recently added to Wikipedia. 
Therefore, for PopQA, we used the December 2020 preprocessed Wikipedia corpus provided by 
\cite{izacard2022few} and generated document embeddings.\footnote{\url{https://github.com/facebookresearch/atlas}}}
The issues of performance variance from different Wikipedia dumps have been reported by prior work~\citep{asai2019learning,izacard2022few}. 
Yet, we observe limited effectiveness of such off-the-shelf retrieval models trained primarily on knowledge-intensive tasks for open-ended generation (e.g., instruction following). 
{Recent or concurrent work studies instruction-tuning of retrieval systems~\citep{asai-etal-2023-task} or joint training of retrieval and LM components~\citep{lin2023radit}, while we leave exploring the effectivess of such appraoches for future work. }
For bio generation and open-domain QA tasks, we additionally retrieve five documents using Google Programmable Search\footnote{\url{https://programmablesearchengine.google.com/about/}} and search documents from English Wikipedia. As this API only provides snippets, we retrieve Wikipedia introductory paragraphs for the corresponding entities. 

\paragraph{Detailed experimental settings for individual datasets.}
For OpenQA datasets, we set the maximum new token number to 100 tokens. For closed-set tasks (PubHealth and ARC-C), we set the maximum new token length to 50 for all baselines. 
For \model inference on PubHealth and ARC-C, instead of determining the output with the highest score~\ref{eq:reward} as in other tasks, we aggregate the scores for each option and select the answer option with the highest score. 
We found in zero-shot settings of fact checking, some LLMs can generate capitalized class labels (e.g., True) while our gold labels are lower-cased. Therefore, across different LMs, for fact checking, we lowercase the predictions. 
In multiple choice tasks, we found some models generate answers in slightly different ways (e.g., (A) instead of A). We slightly modify instructions for each LLM to avoid such format violations, and further conduct string matching between each candidate and model predictions if format violations still remain. After that processing, in closed set tasks, model predictions match one of the gold classes in almost all cases. 
For ALCE, we found that Llama2-chat tend to generate significantly lower outputs than other models (e.g., on average, their output is nearly 100 token, while ChatGPT generates 40 tokens on average), resulting in inflated str-em scores. We limit the maximum generation length to 100 tokens for all baselines to avoid this issue, rather than the original 300 tokens in the ALCE paper. Consequently, all of the baseline output length is within 30-60 tokens. 
For FactScore, we set the maximum new token length to 500 for baselines and 200 for \model at each segment level. 

\paragraph{Task-specific instructions.} Table~\ref{tab:list_of_instructions} shows the list of the instructions used during evaluations. For Open-domain QA, we do not provide explicit instructions. 

\begin{table*}[t!]
\renewcommand{\arraystretch}{1.2}
\setlength{\tabcolsep}{2pt}
\footnotesize
    \centering
    \begin{tabular}{lp{12cm}}
\toprule
\textbf{Dataset} & \textbf{Instruction} \\\midrule
ARC-C & Given four answer candidates, A, B, C and D, choose the best answer choice. Please answer with the capitalized alphabet only, without adding any extra phrase or period. \\
PubHealth & Is the following statement correct or not? Say true if it's correct; otherwise, say false. Don't capitalize or add periods, just say ``true'' or ``false''. \\
Bio Generation & Tell me a bio about \texttt{[Person Name]}\\
ASQA (baseline) & Instruction: Write an accurate, engaging, and concise answer for the given question using only the provided search results (some of which might be irrelevant) and cite them properly. Use an unbiased and journalistic tone. Always cite for any factual claim. When citing several search results, use [1][2][3]. Cite at least one document and at most three documents in each sentence. If multiple documents support the sentence, only cite a minimum sufficient subset of the documents. \\
ASQA (ours) & Answer the following question. The question may be ambiguous and have multiple correct answers, and in that case, you have to provide a long-form answer including all correct answers. \\
\bottomrule
 \end{tabular}
    \caption{Full list of instructions used during zero-shot evaluations. For open-domain QA, we don't use any task specific instruction and simply use the original questions as input query. }\label{tab:list_of_instructions}
\end{table*}

\section{Results}
\subsection{Analysis}
\paragraph{Reliance on parametric- and non-parametric memories.}
We conduct analysis on how frequently model answers come from retrieved passages (non-parametric memories) or their own parametric memories. 
On two open-domain QA datasets, TriviaQA and PopQA, we conduct the following analysis: 1) sample query models successfully answer correctly, 2) for each query in this group, check whether the matched ground-truth answer is a sub-string of the retrieved passage or not. We evaluate \model 7B, Alpaca 7B, Alpaca 13B, and Llama2-Chat-13B. 
We found that \model significantly less frequently generates answers that are not included in the provided evidence; in particular, in Alpaca 30B, 20\% of the correct predictions are not included in the provided passages, followed by Llama2-chat 13B (18\%) and Alpaca (15\%), while it is only 2\% in \model. When retrieved passages are not relevant, \model generates \crel=\texttt{Irrelevant}, indicating that the following answers may not be factually grounded, while those instruction-tuned models continue to generate plausible answers. 

\subsection{Human Evaluation Examples}
Table~\ref{tab:human_evaluations} shows examples with human evaluations on S\&P and correctness of \crel and \cgr reflection tokens. 

\begin{table*}[t!]
\renewcommand{\arraystretch}{1.2}
\setlength{\tabcolsep}{2pt}
\footnotesize
    \centering
    \begin{tabular}{p{14cm}}
\toprule
{\bf Input} What genre is A Question and Answer Guide to Astronomy? (\texttt{PopQA})\\
{\bf Output} [\ret=\texttt{Yes}] \texttt{<p>}{\textcolor{gray}{\bf Reference [1]}}\texttt{<p>}[\crel=\texttt{Relevant}] A Question and Answer Guide to Astronomy is a non-fiction book that falls under the genre of Astronomy and Cosmology. [\cgr=\texttt{Fully Supported}][\cuse=5] \\\hdashline
{\bf Reference [1]} A Question and Answer Guide to Astronomy is a book about astronomy and cosmology, and is intended for a general audience. The book was written by Pierre-Yves Bely, Carol Christian, and Jean-Rene Roy, and published in English by Cambridge University Press in 2010. It was originally written in French. The content within the book is written using a question and answer format. It contains some 250 questions, which The Science Teacher states each are answered with a concise and well-formulated essay that is informative and readable. The Science Teacher review goes on to state that many of the answers given in the book are little gems of science writing. The Science Teacher summarizes by stating that each question is likely to be thought of by a student, and that the answers are informative, well constructed, and thorough. The book covers information about the planets, the Earth, the Universe, practical astronomy, history, and awkward questions such as astronomy in the Bible, UFOs, and aliens. Also covered are subjects such as the Big Bang, comprehension of large numbers, and the Moon illusion. \\\hdashline
{\bf S\&P } Supported and Plausible. \\
{\bf Correctness of \crel and \cgr: } Both \crel and \cgr are correct, as the Reference [1] provides detailed information of the book including the genre, and the model output is fully supported by the reference. \\
{\bf Input }Who is the author of The Lie? (\texttt{PopQA})\\
{\bf Output } [\ret=\texttt{Yes}] \texttt{<p>}{\textcolor{gray}{\bf Reference [1]}}\texttt{<p>}[\crel=\texttt{Relevant}] The author of The Lie is Sam Harris.[\cgr=\texttt{Fully Supported}][\cuse=5] \\\hdashline
{\bf Reference [1]} Lying is a 2011 long-form essay book by American author and neuroscience expert Sam Harris. Harris argues that we can radically simplify our lives and improve society by merely telling the truth in situations where others often lie. \\\hdashline
{\bf S\&P } Plausible but not supported. \\
{\bf Correctness of \crel and \cgr } Both \crel and \cgr are incorrect, as the Reference [1] is about a different book whose title is similar to The Lie. \\\midrule
{\bf Input} Tell me a bio about Ronaldinho (\texttt{Bio})\\
{\bf Output} [\ret=\texttt{Yes}] \texttt{<p>}{\textcolor{gray}{\bf Reference [1]}}\texttt{<p>}[\crel=\texttt{Relevant}] Ronaldinho is a retired professional footballer from Brazil.[\cgr=\texttt{Fully Supported}][\ret=\texttt{Continue}]He played as an attacking midfielder and was known for his technical skills, creativity, dribbling ability, and accuracy from free kicks[\ret=\texttt{Continue}]He won two FIFA World Player of the Year awards and a Ballon d'Or.[\cuse=5] \\\hdashline
{\bf Reference [1]} Ronaldo de Assis Moreira (born 21 March 1980), commonly known as Ronaldinho Gaúcho or simply Ronaldinho, is a Brazilian retired professional footballer who played mostly as an attacking midfielder, but was also deployed as a winger. Considered one of the best players of his generation and regarded by many as one of the greatest of all time, Ronaldinho won two FIFA World Player of the Year awards and a Ballon d'Or. A global icon of the sport, he was renowned for his technical skills, creativity, dribbling ability and accuracy from free-kicks, as well as his use of tricks, feints, no-look passes and overhead... \\\hdashline
{\bf S\&P} Plausible and fully supported. \\
{\bf Correctness of \crel and \cgr }Both \crel and \cgr are correct. \\\midrule
{\bf Input} Tell me a bio about Sergei Bodrov (\texttt{Bio})\\
{\bf Output} [\ret=\texttt{Yes}] \texttt{<p>}{\textcolor{gray}{\bf Reference [1]}}\texttt{<p>}[\crel=\texttt{Relevant}] Sergei Vladimirovich Bodrov (born June 28, 1948) is a Russian film director, screenwriter, and producer.[\cgr=\texttt{Fully Supported}][\ret=\texttt{Continue}]He was born in Moscow, Russia.[\ret=\texttt{Continue}]He is the president of the Russian Guild of Film Producers.[\cuse=5] \\\hdashline
{\bf Reference [1]} Sergei Vladimirovich Bodrov (born June 28, 1948) is a Russian film director, screenwriter, and producer. In 2003 he was the President of the Jury at the 25th Moscow International Film Festival.  \\\hdashline
{\bf S\&P} Plausible but only partially supported. The the Reference [1] does not explicitly mention the person is born in Moscow and there is no description about the Russian Guild of Film Producers. \\
{\bf Correctness of \crel and \cgr } \crel and \cgr are correct as the reference document is highly relevant and the first sentence is fully supported. \\
\bottomrule
 \end{tabular}
    \caption{Examples of human evaluation results.}\label{tab:human_evaluations}
\end{table*}

\subsection{Qualitative Examples}
Table~\ref{tab:examplse_test} shows several examples predicted by our \model (13B). 
The first example is the model output to an ASQA question. The first reference states that Emperor Constantine made Sunday a day of rest from labor, and further the second citation supports the fact that the official adoption of Sunday as a day of rest by Constantine in AD 321. 
In the second example, the model predicts \texttt{Contradictory} to the first output as the output says the person has served as the CEO since 2010, while the passage says he stepped down as CEO in 2015. Indicating those factual contradictions as reflection tokens enables to enforcement of hard control and also verification of model outputs easily. 
In the third example, while the generation is mostly correct, \model predicts \texttt{Partially Support} to the statement listing the name of the songs, as they were not explicitly mentioned. 

\begin{table*}[t!]
\renewcommand{\arraystretch}{1.2}
\setlength{\tabcolsep}{2pt}
\footnotesize
    \centering
    \begin{tabular}{p{14cm}}
\toprule
{\bf Input} When did sunday became the seventh day of the week (\texttt{asqa})\\
{\bf Output} [\ret=\texttt{Yes}] \texttt{<p>}{\textcolor{gray}{\bf Reference [1]}}\texttt{<p>}[\crel=\texttt{Relevant}] In 321 AD, the Roman Emperor Constantine the Great made Sunday a day of rest from labor, stating: ``On the venerable Day of the Sun let the magistrates and people residing in cities rest, and let all workshops be closed''.[\cgr=\texttt{Fully Supported}] [\ret=\texttt{Yes}] \texttt{<p>}{\bf \textcolor{gray}{Reference [2]}}\texttt{<p>}[\crel=\texttt{Relevant}] The official adoption of Sunday as a day of rest by Constantine the Great in AD 321 made Sunday the seventh day of the week.{\bf [2]}[\cgr=\texttt{Fully Supported}][\cuse=5] \\\hdashline
{\bf Reference [1]} Emperor Constantine then enacted the first Sunday Laws, for ``the venerable Day of the Sun'' in 321 A.D. On March 7, 321, the Roman emperor Constantine issued a decree making Sunday a day of rest from labor, stating: We find a good description of this in sources such as the book Rest Days which states: Early Christian observance of both the spiritual seventh-day sabbath and a Lord\'s Day assembly is evidenced in Ignatius\'s letter to the Magnesians ca. 110.  \\
{\bf Reference [2]} Despite the official adoption of Sunday as a day of rest by Constantine, the seven-day week and the nundial cycle continued to be used side-by-side until at least the Calendar of 354 and probably later. ...  The fact that the canon had to be issued at all is an indication that adoption of Constantine's decree of 321 was still not universal ...  \\\midrule
{\bf Input} Tell me a bio about Rory Tapner. (\texttt{bio generation}) \\
{\bf Output} 
\texttt{<p>}{\textcolor{gray}{\bf Reference [1]}}\texttt{<p>}[\crel=\texttt{Relevant}] Rory Tapner is a British businessman who has served as the Chief Executive Officer of Coutts, a private banking and wealth management company, since 2010.[\cgr=\texttt{Contradictory}], {\bf Reference [2]}\texttt{<p>}[\crel=\texttt{Relevant}] Rory Tapner was born on 30 September 1959 in London, England.[\cgr=\texttt{Fully Supported}]\\\hdashline
{\bf Reference [1]} Rory Tapner Rory Tapner  is a UK businessman who, in September 2010 was appointed CEO of Coutts, the Wealth Management division of Royal Bank of Scotland Group. In February 2015 he stepped down as CEO when he was replaced in the role by Alison Rose \\
{\bf Reference [2]} Holdings (from August 2016); and the Council of the University of Buckingham (from July 2017). Rory Tapner Rory Tapner (born 30 September 1959) is a UK businessman ... \\\midrule
{\bf Input} Tell me a bio about G. Venugopa. (\texttt{bio generation}) \\
{\bf Output (ranked 1)} 
[\ret=\texttt{Yes}] \texttt{<p>}{\textcolor{gray}{\bf Reference [1]}}\texttt{<p>}[\crel=\texttt{Relevant}] G. Venugopal is a popular playback singer in the Malayalam film industry. [\ret=\texttt{Continue}] He has sung over 300 songs in Malayalam, Tamil, and Hindi movies.[\cgr=\texttt{Partially Supported}] [\ret=\texttt{Yes}] \texttt{<p>}{\textcolor{gray}{\bf Reference [2]}}\texttt{<p>}G. Venugopal was born on December 10, 1960 in India.[\cgr=\texttt{Fully Supported}] \\\hdashline
{\bf Reference [1]} 'G. Venugopal G Venugopal (born 10 December 1960) is an Indian playback singer known for his work in Malayalam films. He started his singing career in the film ``Odaruthammaava Aalariyaam'' (1984). Since then he has sung in more than 300 films; and has over 500 private albums to his credit. The many awards he has won include the coveted Best Playback Singer (Kerala State) which he won thrice, the Film Critics award and the Kerala Kaumudi Gallop Poll award (twice). He won The Kerala State award for Best Playback singer (2004) third time for the song "aadedee..aadaadedee" from the film "Ullam".Venugopal' \\
{\bf Reference [2]} Kerala State Film Awards: Kerala Film Critics Award Kerala State Government Award (Professional Drama): Asianet Film Awards: Kerala Kaumadi Gallup Poll Award: Academic G. Venugopal G Venugopal (born 10 December 1960) is an Indian playback singer known for his work in Malayalam films.  \\
\bottomrule
 \end{tabular}
    \caption{Examples of outputs.}\label{tab:examplse_test}
\end{table*}

\section{Full List of Instructions and Demonstrations for GPT-4}
\label{sec:instruction_gpt_4}
Here, we show the instructions and demonstrations used to prompt GPT-4 to collect reflection tokens. 
Table~\ref{tab:algo_box} shows the instructions and demonstrations for the initial retrieval token. 
Table~\ref{tab:algo_box_multiple_retrieval} shows the instruction and demonstrations used to collect the three-way output tokens for \ret~given instruction, preceding sentences, and previously retrieved passages. 
Due to the longer demonstration and test input, we only use a single demonstration. 
Table~\ref{tab:algo_box_relevant} shows an instruction and demonstrations used to collect the three-way output tokens for \crel. 
Table~\ref{tab:algo_box_grd} shows an instruction and demonstrations used to collect the three-way output tokens for \crel.
Table~\ref{tab:algo_box_use} shows an instruction and demonstrations used to collect the five-way output tokens for \cuse.
\begin{table*}[h!]
\begin{tcolorbox}
{\bf Instructions}\\ 
Given an instruction, please make a judgment on whether finding some external documents from the web (e.g., Wikipedia) helps to generate a better response. Please answer [Yes] or [No] and write an explanation. 
\tcblower
{\bf Demonstrations}  \\
{\bf Instruction} Give three tips for staying healthy. \\
{\bf Need retrieval?} [Yes] \\
{\bf Explanation} There might be some online sources listing three tips for staying healthy or some reliable sources to explain the effects of different behaviors on health. So retrieving documents is helpful to improve the response to this query. \\ 
\\
{\bf Instruction} Describe a time when you had to make a difficult decision. \\ 
{\bf Need retrieval?} [No] \\ 
{\bf Explanation} This instruction is asking about some personal experience and thus it does not require one to find some external documents. \\
\\
{\bf Instruction} Write a short story in third person narration about a protagonist who has to make an important career decision. \\
{\bf Need retrieval?} [No] \\
{\bf Explanation} This instruction asks us to write a short story, which does not require external evidence to verify. \\
\\
{\bf Instruction} What is the capital of France?\\
{\bf Need retrieval?} [Yes] \\
{\bf Explanation} While the instruction simply asks us to answer the capital of France, which is a widely known fact, retrieving web documents for this question can still help. \\
\\
{\bf Instruction}  Find the area of a circle given its radius. Radius = 4 \\
{\bf Need retrieval?} [No]\\
{\bf Explanation} This is a math question and although we may be able to find some documents describing a formula, it is unlikely to find a document exactly mentioning the answer. \\
\\
{\bf Instruction} Arrange the words in the given sentence to form a grammatically correct sentence. quickly the brown fox jumped\\
{\bf Need retrieval?} [No]\\
{\bf Explanation}  This task doesn't require any external evidence, as it is a simple grammatical question. 
\\
\\
{\bf Instruction} Explain the process of cellular respiration in plants.\\
{\bf Need retrieval?} [Yes]\\
{\bf Explanation} This instruction asks for a detailed description of a scientific concept, and is highly likely that we can find a reliable and useful document to support the response. 
\end{tcolorbox}
\caption{\label{tab:algo_box} Instructions and demonstrations for \ret~aspect given the input only. }
\end{table*} 

\begin{table*}[h!]
\begin{tcolorbox}
{\bf Instructions}\\ 
You will be provided with an instruction, evidence, output sentence, and preceding sentences (optional). If the preceding sentence is given, the output should be the sentence that follows those preceding sentences.  Your task is to determine whether the information in the output sentence can be fully verified by the evidence or if it requires further external verification. There are three cases: \\
- If the output sentence can be verified solely with the evidence, then respond with [Continue to Use Evidence]. \\
- If the sentence doesn’t require any factual verification (e.g., a subjective sentence or a sentence about common sense), then respond with  [No Retrieval]. \\
- If additional information is needed to verify the output sentence, respond with [Retrieval]. Please provide explanations for your judgments. 
\tcblower
{\bf Instruction} Explain the use of word embeddings in Natural Language Processing. \\
{\bf Preceding sentences} Word embeddings are one of the most powerful tools available for Natural Language Processing (NLP). They are mathematical representations of words or phrases in a vector space, allowing similarities between words and the context in which they are used to be measured. \\
{\bf Evidence}: 
Word embedding \\
Word embedding is the collective name for a set of language modeling and feature learning techniques in natural language processing (NLP) where words or phrases from the vocabulary are mapped to vectors of real numbers. Conceptually it involves a mathematical embedding from a space with one dimension per word to a continuous vector space with a much lower dimension. 
Output: Word embeddings are useful for tasks such as sentiment analysis, text classification, predicting the next word in a sequence, and understanding synonyms and analogies. \\
{\bf Rating} [Retrieval] \\
{\bf Explanation} The output discusses the applications of word embeddings, while the evidence only discusses the definitions of word embeddings and how they work. Therefore, we need to retrieve other evidence to verify whether the output is correct or not. 
\end{tcolorbox}
\caption{\label{tab:algo_box_multiple_retrieval} Instructions and demonstrations for \ret~aspect given the input, preceding generations, and retrieved passages. }
\end{table*}

\begin{table*}[h!]
\begin{tcolorbox}
{\bf Instructions}\\ 
You'll be provided with an instruction, along with evidence and possibly some preceding sentences. When there are preceding sentences, your focus should be on the sentence that comes after them. Your job is to determine if the evidence is relevant to the initial instruction and the preceding context, and provides useful information to complete the task described in the instruction. If the evidence meets this requirement, respond with [Relevant]; otherwise, generate [Irrelevant].
\tcblower
{\bf Instruction} Given four answer options, A, B, C, and D, choose the best answer. \\
{\bf Input} Earth's rotating causes \\
A: the cycling of AM and PM \\
B: the creation of volcanic eruptions \\
C: the cycling of the tides \\
D: the creation of gravity \\ 
{\bf Evidence} Rotation causes the day-night cycle which also creates a corresponding cycle of temperature and humidity creates a corresponding cycle of temperature and humidity. Sea level rises and falls twice a day as the earth rotates. \\
{\bf Rating} [Relevant] \\
{\bf Explanation} The evidence explicitly mentions that the rotation causes a day-night cycle, as described in the answer option A. \\
\\
{\bf Instruction} age to run for US House of Representatives \\
{\bf Evidence} The Constitution sets three qualifications for service in the U.S. Senate: age (at least thirty years of age); U.S. citizenship (at least nine years); and residency in the state a senator represents at the time of election. \\
{\bf Rating} [Irrelevant] \\
{\bf Explanation} The evidence only discusses the ages to run for the US Senate, not for the House of Representatives.  
\end{tcolorbox}
\caption{\label{tab:algo_box_relevant} Instructions and demonstrations for \crel~aspect given the input only. }
\end{table*}

\begin{table*}[h!]
\begin{tcolorbox}
{\bf Instructions}\\ 
You will receive an instruction, evidence, and output, and optional preceding sentences.  If the preceding sentence is given, the output should be the sentence that follows those preceding sentences. Your task is to evaluate if the output is fully supported by the information provided in the evidence.\\
Use the following entailment scale to generate a score:\\
- [Fully supported] - All information in output is supported by the evidence, or extractions from the evidence. This is only applicable when the output and part of the evidence are almost identical. \\
- [Partially supported] - The output is supported by the evidence to some extent, but there is major information in the output that is not discussed in the evidence. For example, if an instruction asks about two concepts and the evidence only discusses either of them, it should be considered a [Partially supported]. \\
- [No support / Contradictory] - The output completely ignores evidence, is unrelated to the evidence, or contradicts the evidence. This can also happen if the evidence is irrelevant to the instruction.\\\\
Make sure to not use any external information/knowledge to judge whether the output is true or not. Only check whether the output is supported by the evidence, and not whether the output follows the instructions or not.
\tcblower
{\bf Instruction} Explain the use of word embeddings in Natural Language Processing. \\
{\bf Preceding sentences} Word embeddings are one of the most powerful tools available for Natural Language Processing (NLP). They are mathematical representations of words or phrases in a vector space, allowing similarities between words and the context in which they are used to be measured. \\
{\bf Output} Word embeddings are useful for tasks such as sentiment analysis, text classification, predicting the next word in a sequence, and understanding synonyms and analogies. \\
{\bf Evidence} 
Word embedding \\
Word embedding is the collective name for a set of language modeling and feature learning techniques in natural language processing (NLP) where words or phrases from the vocabulary are mapped to vectors of real numbers. Conceptually it involves a mathematical embedding from a space with one dimension per word to a continuous vector space with a much lower dimension. Methods to generate this mapping include neural networks, dimensionality reduction on the word co-occurrence matrix, probabilistic models, explainable knowledge base method, and explicit representation in terms of the context in which words appear. Word and phrase embeddings, when used as the underlying input representation, have been shown to boost the performance in NLP tasks such as syntactic parsing, sentiment analysis, next token predictions as well and analogy detection. \\
{\bf Score} [Fully supported] \\
{\bf Explanation} The output sentence discusses the application of word embeddings, and the evidence mentions all of the applications syntactic parsing, sentiment analysis, next token predictions as well as analogy detection as the applications. Therefore, the score should be [Fully supported].
\end{tcolorbox}
\caption{\label{tab:algo_box_grd} Instructions and demonstrations for \cgr tokens. }
\end{table*}

\begin{table*}[h!]
\begin{tcolorbox}
{\bf Instructions}\\ 
Given an instruction and an output, rate whether the response appears to be a helpful and informative answer to the query, from 1 (lowest) - 5 (highest). We call this score perceived utility. The detailed criterion is as follows: 
5: The response provides a complete, highly detailed, and informative response to the query, fully satisfying the information needs. 
4: The response mostly fulfills the need in the query, while there can be some minor improvements such as discussing more detailed information, having better structure of the response, or improving coherence.   
3: The response is acceptable, but some major additions or improvements are needed to satisfy users’ needs. 
2: The response still addresses the main request, but it is not complete or not relevant to the query. 
1: The response is barely on-topic or completely irrelevant. 
\tcblower
{\bf Instruction} Who is the current prime minister of the UK as of 2023? \\
{\bf Output} Boris Johnson was the prime minister of the UK from 2019 - 2022. \\
{\bf Perceived utility} 2 \\
{\bf Explanation} While the output provides a factually correct statement about the UK prime minister from 2019 to 2022, this instruction asks who the prime minister is as of 2023, so it doesn't answer the instruction. Therefore, the utility is 2. \\
\\
{\bf Instruction} Given a description of a travel destination, recommend 10 tourist attractions with detailed descriptions of each. The travel destination is Tokyo, Japan.\\
{\bf Output} 
Tokyo is a vibrant city full of exciting tourist attractions. Some must-see sights include the Tokyo Skytree, Tokyo Disneyland, Sensoji Temple, Meiji Shrine, Tsukiji Fish Market, Harajuku, and Shinjuku Gyoen. \\
{\bf Perceived utility} 3 \\
{\bf Explanation} This output doesn't provide descriptions of each attraction and the number of the attractions is also less than 10. While this output partially answers the instructions, it doesn't match the instructions strictly. 
\end{tcolorbox}
\caption{\label{tab:algo_box_use} Instructions and demonstrations for \cuse tokens. }
\end{table*}